\documentclass{article}

    \PassOptionsToPackage{numbers, compress}{natbib}
     \usepackage[preprint]{neurips_2025}

\usepackage[utf8]{inputenc} % allow utf-8 input
\usepackage[T1]{fontenc}    % use 8-bit T1 fonts
\usepackage{hyperref}       % hyperlinks
\usepackage{url}            % simple URL typesetting
\usepackage{booktabs}       % professional-quality tables
\usepackage{amsfonts}       % blackboard math symbols
\usepackage{nicefrac}       % compact symbols for 1/2, etc.
\usepackage{microtype}      % microtypography
\usepackage{xcolor}         % colors

\usepackage{bm}
\usepackage{amsmath}
\usepackage{algorithm,algorithmic}
\usepackage{graphicx}
\usepackage{caption}
\usepackage{subcaption}
\graphicspath{{images/}}
\graphicspath{{Sections/}}
\usepackage{multirow}
\usepackage{longtable}
\usepackage{dirtree}
\usepackage{pdfpages}
\usepackage{amssymb}
\usepackage{enumitem}
\usepackage{wrapfig}
\usepackage{nicefrac}
\usepackage{tcolorbox}
\tcbuselibrary{listings,breakable,skins}   % ← added “breakable”

\title{Backpropagation-Free Metropolis-Adjusted\\ Langevin Algorithm}

\author{%
	Adam D. Cobb,  \;\; 
	Susmit Jha 
	\\
	\\
	\enspace Computer Science Laboratory, SRI
}

\begin{document}

\maketitle

\begin{abstract}
  Recent work on backpropagation-free learning has shown that it is possible to use forward-mode automatic differentiation (AD) to perform optimization on differentiable models. Forward-mode AD requires sampling a tangent vector for each forward pass of a model. The result is the model evaluation with the directional derivative along the tangent. In this paper, we illustrate how the sampling of this tangent vector can be incorporated into the proposal mechanism for the Metropolis-Adjusted Langevin Algorithm (MALA). As such, we are the first to introduce a backpropagation-free gradient-based Markov chain Monte Carlo (MCMC) algorithm. We also extend to a novel backpropagation-free position-specific preconditioned forward-mode MALA that leverages Hessian information. Overall, we propose four new algorithms: Forward MALA; Line Forward MALA; Pre-conditioned Forward MALA, and Pre-conditioned Line Forward MALA. We highlight the reduced computational cost of the forward-mode samplers and show that forward-mode is competitive with the original MALA, while even outperforming it depending on the probabilistic model. We include Bayesian inference results on a range of probabilistic models, including hierarchical distributions and Bayesian neural networks.
\end{abstract}

\section{Introduction}\label{sec:intro}
% Importance of gradient-based MCMC in Bayesian inference
Gradient-based Markov chain Monte Carlo (MCMC) approaches are often the sampling algorithm of choice when it comes to performing Bayesian inference over differentiable probabilistic models. Gradient evaluations improve the ability of these algorithms to scale with dimension and achieve faster convergence to the target distribution \citep{neal2011mcmc}. As an example, Bayesian neural networks (BNNs) are a model class that favors gradient-based MCMC algorithms for sampling. Therefore there is significant value in developing new gradient-based MCMC algorithms, especially if new approaches can reduce the cost of evaluating gradients.

% Introduce Forward-Mode and why it is interesting
To evaluate gradients in large hierarchical (or deep) models, machine learning pipelines almost exclusively rely on reverse-mode AD (aka backpropagation). This requires a forward pass which stores the intermediate values, followed by a backward pass that collects these values to evaluate the full gradient. On the other hand, forward-mode AD implements the chain rule in the forward direction. For a function $f(\bm{\theta})$, where $\bm{\theta}\in\mathbb{R}^D$, it requires setting a tangent vector, $\mathbf{v}\in\mathbb{R}^D$, of the same dimension. Then the resulting forward pass uses the tangent vector to evaluate the Jacobian vector product (JVP). For a model with a single valued output (e.g. log-likelihood) the JVP corresponds to the directional derivative in the direction of the tangent vector, $\nabla f(\bm{\theta})\cdot \mathbf{v}$. Recent work by \citet{baydin2022gradients} has shown that sampling the tangent vector from an IID Gaussian distribution results in an unbiased estimate of the gradient in a single forward pass. %\textcolor{red}
If we can leverage these estimates of the gradient instead of using a full reverse-mode step then there are two potential advantages: (1) The memory footprint of forward-mode is less than backpropagation since there is no required storage for the reverse pass; (2) The runtime cost of a forward evaluation is approximately only twice that of a single function call \citep{griewank2008evaluating}. 

In this paper we connect forward-mode AD with gradient-based MCMC for Bayesian inference. 
Our key insight comes from incorporating the distribution over the tangent vectors into the proposal mechanism for our new forward-mode-only Metropolis-adjusted Langevin algorithm. We introduce two types of samplers: (1) a single-stage forward-mode sampler (FMALA); and (2) a two-stage forward-mode sampler with a Gibbs-style step to sample a line
followed by the forward-mode MALA step along that line (Line-FMALA). Furthermore, we define two variants of these samplers, one that uses first-order information and the other that uses Hessian information in the form of a second-order forward-mode step. The latter approach provides the advantages of preconditioning each update step with position-specific curvature information. As a result, in this paper we propose the following forward-mode samplers: (1) FMALA; (2) Line-FMALA; (3) PC-FMALA; (4) PC-Line-FMALA. We compare all four variants with original MALA across multiple models. 
Overall this paper has the following contributions:
% {
% \setlength{\topsep}{0pt} % Reduces the space above the list
% \setlength{\partopsep}{0pt} % Reduces the space below the list
\begin{itemize}[leftmargin=*, topsep=0pt]
    \item We are the first to introduce backpropagation-free gradient-based MCMC.
    \item We define four new sampling schemes, including a novel two-stage line-based sampling scheme and the use of second-order forward-mode AD to precondition FMALA.
    \item We show the performance of forward-mode MALA approaches are competitive with the original reverse-mode MALA and can even outperform MALA depending on the probabilistic model. This is significant because the runtime-cost and memory-cost of forward-mode approaches are substantially lower than the reverse-mode counterparts. 
    \item We include experimental results on a range of probabilistic models including hierarchical distributions and BNNs.
\end{itemize}

%Structure
The rest of the paper is structured as follows. Section \ref{sec:rel} includes related work and then Section \ref{sec:prel} provides background on forward-mode AD and MALA. Section \ref{sec:method}, introduces our four new forward-mode samplers. Section \ref{sec:exp} contains our experiments and then we conclude in Section \ref{sec:conc}. Code is available at \url{https://github.com/SRI-CSL/fojax}.

\section{Related work}\label{sec:rel}
% Forward-mode in optimization
While there has been no previous work on the use of forward-mode AD in MCMC, there has been a recent interest in the use of forward-mode AD for backpropagation-free learning mechanisms \citep{silver2021learning,baydin2022gradients,hinton2022forward,fournier2023can, cobb2024second}. The interest in moving away from backpropagation comes from two main motivations. First, it is presumed unlikely that biological systems follow a reverse-mode AD learning paradigm \citep{bengio2015towards, hinton2022forward}. Second, reverse-mode AD comes with certain architectural constraints and requirements. For example, \citet{jaderberg2017decoupled} refer to the problem of the backward lock, meaning that parameters cannot be updated until all the dependent parameters have experienced both the forward and backward pass. 
Therefore it is vital to \textbf{explore cheaper alternatives} that \textbf{reduce training and energy costs}. Forward-mode AD is known to be a cheaper operation compared to reverse-mode AD.

\citet{baydin2022gradients} introduced Forward Gradient Descent (FGD), which relies on forward-mode AD to estimate the gradients in an optimization routine. This built on previous work in weight perturbation approaches \cite{pearlmutter1994fast}. Since the gradient estimator is inherently noisy, FGD suffers from increasing variance with parameter dimension. As such, work on reducing this variance has included incorporating local reverse-mode steps (or local losses) \citep{ren2022scaling}, as well as improving tangent guesses \citep{fournier2023can}. Another recently explored direction is to leverage second-order forward-mode operations to perform second-order optimization, which has also shown promise \citep{cobb2024second}. We will also use second-order forward-mode steps within this paper, but we will use it to build position-specific metrics for MCMC.

Finally, gradient-based MCMC approaches are a proven algorithm of choice when performing Bayesian inference over high-dimensional differentiable models \citep{duane1987hybrid, neal1995bayesian}. They have favorable scaling with dimension \citep{neal2011mcmc} and can achieve state-of-the-art in uncertainty quantification benchmarks \citep{izmailov2021bayesian}. One common challenge in applying these sampling schemes, such as for MALA and Hamiltonian Monte Carlo (HMC), is in tuning the step size. In the literature this is often tackled by adapting the effective step-size directly \citep{hoffman2014no}, or integrating local geometry into the sampler \citep{girolami2011riemann, betancourt2013general}.

\section{Preliminaries}\label{sec:prel}

Throughout the paper we define $\bm{\theta}\in\mathbb{R}^D$ to be the parameters of a distribution, $p(\bm{\theta})$. We also define $f(\bm{\theta}) = \log p(\bm{\theta})$. 

\subsection{Forward-mode automatic differentiation}\label{sec:AD}
Forward-mode automatic differentiation \citep{wengert1964simple} applies the chain rule in the \textit{forward} direction. It requires setting a tangent vector, $\mathbf{v}\in\mathbb{R}^D$, prior to each function call, $f(\bm{\theta})$. Each forward-mode evaluation provides the function value at $\bm{\theta}$, as well as the Jacobian vector product (JVP). For a unidimensional output (as is often the case for machine learning models), the JVP is the directional derivative,  $\nabla f(\bm{\theta}) \cdot \mathbf{v}$. To implement forward-mode AD, one generally uses dual numbers. These are available in most AD libraries such as PyTorch \citep{paszke2019pytorch} and JAX \citep{jax2018github}. Since dual numbers act to truncate a Taylor series to the first-order, one can also extend dual numbers to evaluate higher-order terms of the Taylor series, such as the second order quadratic term, $\mathbf{v}^{\top} \nabla^2 f(\bm{\theta})\mathbf{v}$. The result are the forward-mode evaluations for first-order, 
\begin{equation}\label{eq:fir-order}
	F_1(\bm{\theta}, \mathbf{v}) \rightarrow [f(\bm{\theta}), \nabla f(\bm{\theta}) \cdot \mathbf{v}],
\end{equation}
and second order,
\begin{equation}\label{eq:sec-order}
	F_2(\bm{\theta}, \mathbf{v}) \rightarrow [f(\bm{\theta}), \nabla f(\bm{\theta}) \cdot \mathbf{v}, \mathbf{v}^{\top} \nabla^2 f(\bm{\theta})\mathbf{v}].
\end{equation}
Unlike reverse-mode AD, forward-mode AD does not require storing the intermediate values of the computation graph since there is no backward pass. As a result the theoretical memory cost of implementing forward-mode AD is approximately twice that of a single function evaluation. 
% as one only needs to compute the JVPs and the values of the intermediate functions en route to the final function evaluation and JVP. 
The time complexity of a forward pass is also approximately twice that of a single function call due to similar reasoning \citep{griewank2008evaluating}. These values may vary in practice depending on the implementation. Finally, to form the basis of forward-mode optimizers such as FGD, one samples tangent vectors according to $\mathbf{v}\sim \mathcal{N}(\mathbf{0}, \mathbf{I})$, which gives the unbiased estimate of the gradient, $\mathbb{E}[(\nabla f(\bm{\theta}) \cdot \mathbf{v}) \mathbf{v}] = \nabla f(\bm{\theta})$.

\subsection{Metropolis-adjusted Langevin algorithm}\label{sec:MALA}
The Metropolis-adjusted Langevin algorithm \citep{rossky1978brownian, roberts2002langevin} is based on the first-order Euler discretization of the Langevin diffusion equation with stationary distribution, $p(\bm{\theta})$. This is followed by the Metropolis-Hastings (MH) update step. The procedure for proposing a new $\bm{\theta}_*$ conditional on $\bm{\theta}_t$ is given by
\begin{equation}\label{eq:MALA}
	\bm{\theta}_* = \bm{\theta}_t + \frac{\eta^2}{2} \nabla \log p( \bm{\theta}_t) + \eta \mathbf{z}_t,
\end{equation}
where $\mathbf{z}_t \sim \mathcal{N}(\mathbf{0}, \mathbf{I})$, and $\eta$ is the integration step size. Equation \eqref{eq:MALA} defines the proposal distribution, $q(\bm{\theta}_*|\bm{\theta}_t) = \mathcal{N}(\bm{\mu}(\bm{\theta}_t, \eta), \eta^2 \mathbf{I})$, where $\bm{\mu}(\bm{\theta}_t, \eta) = \bm{\theta}_t + \frac{\eta^2}{2} \nabla \log p( \bm{\theta}_t)$. The MH acceptance log-probability is given by $\gamma = \min\{0, \log p(\bm{\theta}_*) + \log q(\bm{\theta}_t|\bm{\theta}_*) - \log p(\bm{\theta}_t)- \log q(\bm{\theta}_*|\bm{\theta}_t) \}$. The full algorithm samples a new $\bm{\theta}_*$ and then determines whether to accept and set $\bm{\theta}_{t+1} = \bm{\theta}_*$ according the the MH acceptance probability.

One can extend Equation \eqref{eq:MALA} to account for local geometry by building a position-specific preconditioned MALA proposal using $\mathbf{G}(\bm{\theta})$ as the position-specific metric \citep{girolami2011riemann}:\footnote{We note that \citet{xifara2014langevin} propose a new position-dependent MALA which includes additional third-order derivative information. Here, we limit ourselves to samplers leveraging derivatives up to the second-order.}
\begin{equation}\label{eq:PMALA}
    \bm{\theta}_* = \bm{\theta}_t + \frac{\eta^2}{2} \mathbf{G}^{-1}(\bm{\theta}) \nabla f( \bm{\theta}_t) + \eta \sqrt{\mathbf{G}^{-1}(\bm{\theta}_t)} \mathbf{z}.
\end{equation}
The proposal distribution is now given by $q(\bm{\theta}_*|\bm{\theta}_t) = \mathcal{N}(\bm{\theta}_*; \bm{\mu}_{\mathrm{M}}(\bm{\theta}_t, \eta), \eta^2 \mathbf{G}^{-1}(\bm{\theta}_t))$, where $\bm{\mu}(\bm{\theta}_t, \eta) = \bm{\theta}_t + \frac{\eta^2}{2} \mathbf{G}^{-1}(\bm{\theta}_t) \nabla f( \bm{\theta}_t)$, with the same acceptance probability given by
$
\min \left(1, p(\bm{\theta}_*)q(\bm{\theta}_t|\bm{\theta}_*)/p(\bm{\theta}_t)q(\bm{\theta}_*|\bm{\theta}_t) \right)$.

To determine the metric, \citet{girolami2011riemann}, note that the parameter space of a statistical model is a Riemannian manifold. This can be seen from a first order expansion of the symmetric Kullback--Leibler divergence between two densities, $\mathrm{D}_{\mathrm{S}}(p(\bm{\theta}+\delta \bm{\theta}|\mathbf{x})||p(\bm{\theta}|\mathbf{x})) \approx \delta \bm{\theta} \mathbb{E}_{\mathbf{x}}[\nabla^2_{\bm{\theta}} \log p(\bm{\theta}|\mathbf{x})]\delta \bm{\theta}$, where $\mathbf{x}$ is data. This results in the Fisher-Rao metric, $\mathbf{G}(\bm{\theta}) = \mathbb{E}_{\mathbf{x}}[\nabla^2_{\bm{\theta}} \log p(\bm{\theta}|\mathbf{x})]$ \cite{amari2000methods}. While the metric is positive semi-definite, in practice integrating out $\mathbf{x}$ is often infeasible and as a result people have come up with Hessian-inspired metrics that are well behaved (E.g. see \citet{betancourt2013general}).

\section{Forward-mode MALA}\label{sec:method}

In this section we introduce four new forward-mode MALAs. We first introduce the simplest version of FMALA, and then introduce our Line-FMALA sampler. Thereafter we extend both algorithms to a position-specific pre-conditioned MALA, which uses second-order forward-mode information to apply position-specific conditioning. For all forward-mode MALAs, we sample tangent vectors, $\mathbf{v}$, to leverage both gradient and curvature information via Equations \eqref{eq:fir-order} and \eqref{eq:sec-order}. In this paper, we define a unit vector $\hat{\mathbf{v}} = \frac{\mathbf{v}}{||\mathbf{v}||}$, where $\mathbf{v}\sim \mathcal{N}(\mathbf{0}, \mathbf{I})$. Therefore, $\hat{\mathbf{v}}$ is distributed according to a uniform distribution on the unit sphere $S^{D-1}$ such that $\hat{\mathbf{v}} \sim \mathrm{Uniform}(S^{D-1})$.

\subsection{Forward MALA}

Our first iteration of forward-mode MALA incorporates the first-order forward-mode operator directly into the original MALA proposal mechanism by sampling $\hat{\mathbf{v}}$ and following the update step given by
\begin{equation}\label{eq:FMALA}
	\bm{\theta}_* = \bm{\theta}_t + \frac{\eta^2}{2} \left(\nabla f( \bm{\theta}_t) \cdot \hat{\mathbf{v}}_t\right)\hat{\mathbf{v}}_t + \eta \mathbf{z}_t.
\end{equation}
The proposal distribution now depends on the tangent direction and is written as $q(\bm{\theta}_*|\bm{\theta}_t, \hat{\mathbf{v}}_t) q(\hat{\mathbf{v}}_t) = \mathcal{N}(\bm{\theta}_*; \bm{\mu}_{\mathrm{t,FM}}(\bm{\theta}_t, \hat{\mathbf{v}}_t, \eta), \eta^2 \mathbf{I}) \mathcal{U}(\hat{\mathbf{v}}_t; S^{D-1})$, with the corresponding $\bm{\mu}_{\mathrm{t,FM}}(\bm{\theta}_t,  \hat{\mathbf{v}}_t, \eta) = \bm{\theta}_t + \left(\frac{\eta^2}{2} \nabla f( \bm{\theta}_t) \cdot \hat{\mathbf{v}}_t\right)\hat{\mathbf{v}}_t$. For the proposal defined in the reverse direction, $q(\bm{\theta}_t|\bm{\theta}_*, \hat{\mathbf{v}}_*) q(\hat{\mathbf{v}}_*)$, the Gaussian distribution follows the same form as the forward proposal, except with mean, $\bm{\mu}_{\mathrm{*,FM}}(\bm{\theta}_*,  \hat{\mathbf{v}}_*, \eta) = \bm{\theta}_* + \left(\frac{\eta^2}{2} \nabla f( \bm{\theta}_*) \cdot \hat{\mathbf{v}}_*\right)\hat{\mathbf{v}}_*$. 

\begin{figure}[h!]
  \centering
  \begin{minipage}[t]{0.49\textwidth}
    \vspace{0pt} % Align top
    \centering
    \includegraphics[width=1.07\linewidth]{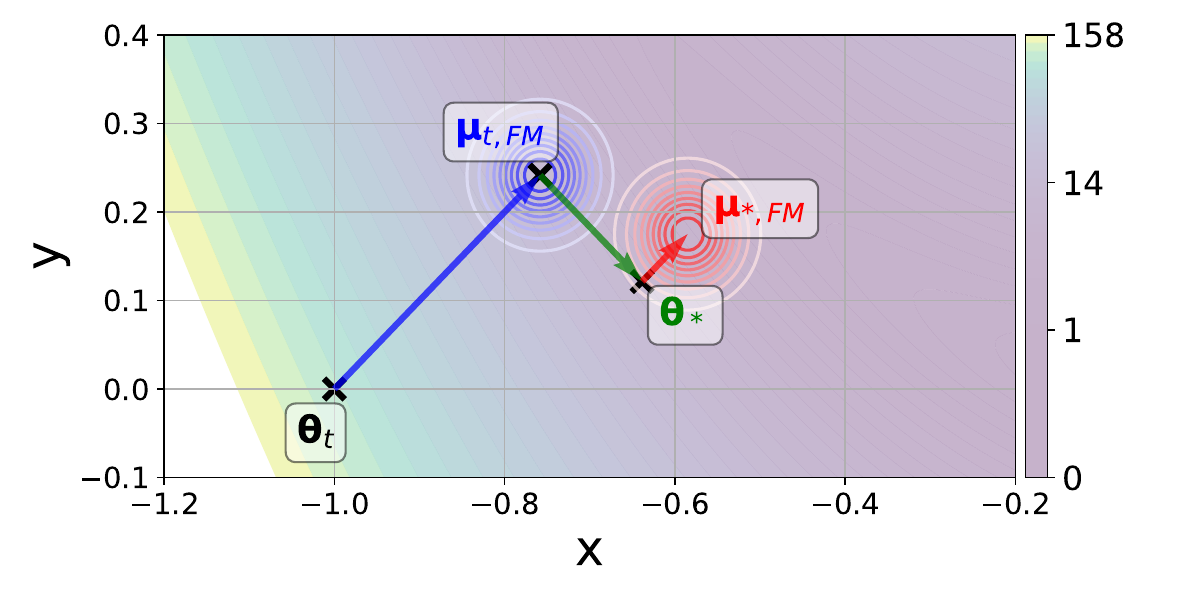}
    \caption{FMALA single update step.}
    \label{fig:fmala_update}
  \end{minipage}
  \hfill
  \begin{minipage}[t]{0.49\textwidth}
    \vspace{0pt} % Align top
    \centering
    \includegraphics[width=1.07\linewidth]{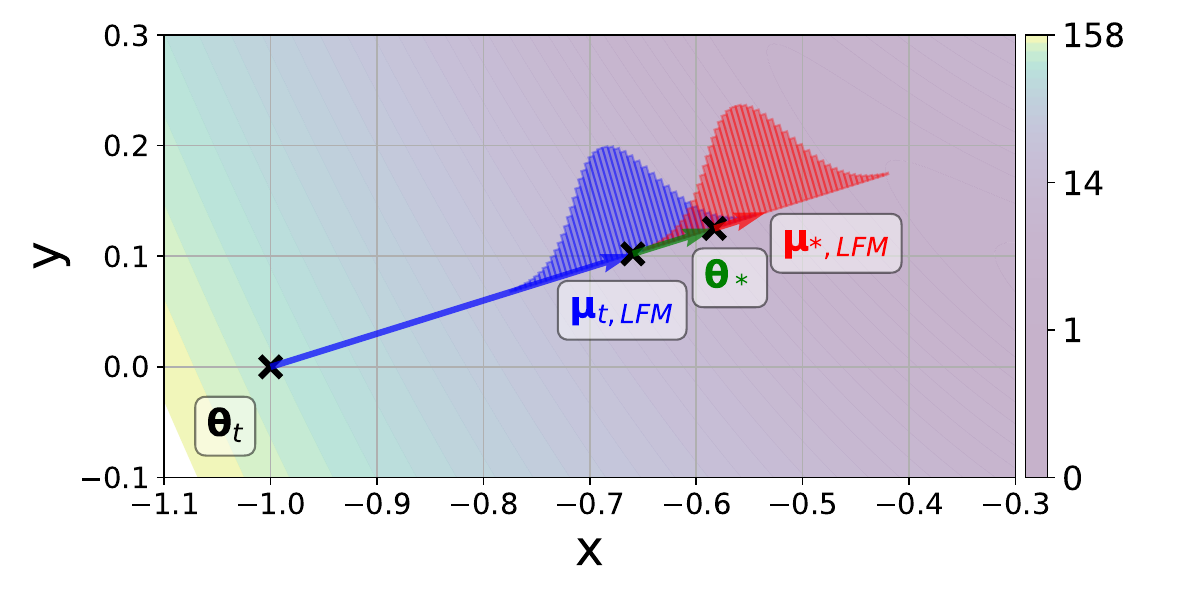}
    \caption{Line-FMALA single update step.}
    \label{fig:line_fmala_update}
  \end{minipage}
\end{figure}

To evaluate $\bm{\mu}_{\mathrm{*,FM}}(\bm{\theta}_*,  \hat{\mathbf{v}}_*, \eta)$ we sample a new $\hat{\mathbf{v}}_*$ and apply $F_1(\bm{\theta}_*, \hat{\mathbf{v}}_*)$. Finally, when building the MH acceptance ratio, both $q(\hat{\mathbf{v}}_*)$ and $q(\hat{\mathbf{v}}_t)$ do not contribute (by design), since all samples from the uniform distribution on the unit sphere are equally likely, giving $q(\hat{\mathbf{v}}_*)=q(\hat{\mathbf{v}}_t)$, and therefore these terms cancel out. Figure \ref{fig:fmala_update} illustrates a single update step within the 2D Rosenbrock function \citep{rosenbrock1960automatic}. We superimpose the contours of both $q(\bm{\theta}_*|\bm{\theta}_t, \hat{\mathbf{v}}_t)$ (blue) and $q(\bm{\theta}_t|\bm{\theta}_*, \hat{\mathbf{v}}_*)$ (red).

\subsection{Line forward MALA}

For our second iteration of forward-mode MALA, we leverage the sampled tangent vector to constrain the update direction. The result is our new forward-mode line MALA (Line-FMALA) approach, which has two stages. Stage one samples a direction (or line) in the form of the tangent vector $\hat{\mathbf{v}}$. Stage two performs a MH update step conditioned on the $\hat{\mathbf{v}}$ direction. The result is to sample from the joint density of $p(\bm{\theta}, \hat{\mathbf{v}})$, where we are only interested in the marginal $p(\bm{\theta})$. 
% This setup has parallels to other gradient-based MCMC samplers, such as for HMC which samples from the joint density over the parameter and momentum variables. 

Using the sampled unit vector $\hat{\mathbf{v}}$ as the update direction, the new Line-FMALA proposal mechanism is given by  
\begin{equation}\label{eq:line_FMALA}
	\bm{\theta}_* = \bm{\theta}_t + \left(\frac{\eta^2}{2} \nabla f( \bm{\theta}_t) \cdot \hat{\mathbf{v}} + \eta z_t\right)\hat{\mathbf{v}},
\end{equation}
where $z_t \in \mathbb{R} \sim \mathcal{N}(0,1)$. This simplifies the proposal distribution by reducing it to sampling a scalar along the sampled tangent vector. Therefore, we rewrite Equation \eqref{eq:line_FMALA} as a scalar update,
\begin{equation}\label{eq:line_FMALA_reduced}
	\bm{\theta}_* \cdot \hat{\mathbf{v}} = \bm{\theta}_t \cdot \hat{\mathbf{v}} + \frac{\eta^2}{2} \nabla f( \bm{\theta}_t) \cdot \hat{\mathbf{v}} + \eta z_t.
\end{equation}
We define $\alpha_* = \bm{\theta}_* \cdot \hat{\mathbf{v}}$ and $\alpha_t = \bm{\theta}_t \cdot \hat{\mathbf{v}}$ to give the reduced proposal distribution $q(\alpha_*|\alpha_t, \hat{\mathbf{v}}) = \mathcal{N}(\alpha_*; \mu_{\mathrm{t,LFM}}(\alpha_t, \hat{\mathbf{v}}, \eta), \eta^2 \mathbf{I})$, 
where $\mu_{\mathrm{t,LFM}}(\alpha_t, \hat{\mathbf{v}}, \eta) = \alpha_t + \frac{\eta^2}{2} \nabla f( \bm{\theta}_t) \cdot \hat{\mathbf{v}}$. For the reverse proposal, $q(\alpha_t|\alpha_*, \hat{\mathbf{v}})$, the mean is now given as $\mu_{\mathrm{*,LFM}}(\alpha_*, \hat{\mathbf{v}}, \eta) = \alpha_* + \frac{\eta^2}{2} \nabla f( \bm{\theta}_*) \cdot \hat{\mathbf{v}}$. A nice behavior of this algorithm is that it only requires sampling an additional scalar value, compared to FMALA, which must sample an additional $\hat{\mathbf{v}}_* \in \mathbb{R}^D$. Figure \ref{fig:line_fmala_update} illustrates the Line-FMALA update step for the same 2D Rosenbrock function. Note that $q(\alpha_*|\alpha_t, \hat{\mathbf{v}})$ (blue) and $q(\alpha_t|\alpha_*, \hat{\mathbf{v}})$ (red) are now 1D Gaussians with densities along the line.

\subsection{Position-specific pre-conditioned forward-mode MALA}

As is often the case with MCMC sampling algorithms, tuning the proposal distribution is a challenge. Specifically, for gradient-based sampling approaches such as MALA this challenge often arises when setting the step size. To exploit these geometric concepts, we leverage the position specific pre-conditioned MALA proposal (as in \citet{girolami2011riemann}, see Section \ref{sec:MALA}), and extend our FMALA and Line-FMALA algorithms to this setting.

To define our pre-conditioned forward-mode MALA (PC-FMALA), we now use second-order information using second-order forward-mode AD as highlighted in Section \ref{sec:AD}. This is a novel way of using all the terms of the second-order forward-mode step to our advantage.  

\subsubsection{Pre-conditioned forward-mode MALA}
We start from the proposal mechanism for position specific pre-conditioned MALA from Equation \eqref{eq:PMALA} and
use the second-order forward pass operation from Equation \eqref{eq:sec-order}. We now build our corresponding new pre-conditioned forward-mode MALA (PC-FMALA) proposal mechanism:
\begin{align}\label{eq:PreCon-FMALA}
	\bm{\theta}_* = \bm{\theta}_t + \frac{\eta^2}{2 |\hat{\mathbf{v}}_t^{\top} \nabla^2 f(\bm{\theta}_t) \hat{\mathbf{v}}_t|} \left(\nabla f( \bm{\theta}_t) \cdot \hat{\mathbf{v}}_t\right)\hat{\mathbf{v}}_t + \eta \sqrt{\left(|\hat{\mathbf{v}}_t^{\top} \nabla^2 f(\bm{\theta}_t) \hat{\mathbf{v}}_t|\right)^{-1}}\mathbf{z}_t.
\end{align}
At this point we highlight that terms $\nabla f( \bm{\theta}_t) \cdot \hat{\mathbf{v}}$ and $\hat{\mathbf{v}}^{\top} \nabla^2 f(\bm{\theta}_t) \hat{\mathbf{v}}$ simply arise from the second-order forward-mode evaluation of $F_2(\bm{\theta}_t, \hat{\mathbf{v}}_t)$. To complete the PC-FMALA sampler, we require the evaluation of both the forward, $q(\bm{\theta}_*|\bm{\theta}_t, \hat{\mathbf{v}}_t)q(\hat{\mathbf{v}}_t)$, and reverse,  $q(\bm{\theta}_t|\bm{\theta}_*, \hat{\mathbf{v}}_*)q(\hat{\mathbf{v}}_*)$, proposal probabilities for the MH acceptance ratio. Each conditional distribution is a Gaussian as before, 
$$q(\bm{\theta}_*|\bm{\theta}_t, \hat{\mathbf{v}}_t) = \mathcal{N}\left(\bm{\theta}_*; \bm{\mu}_{\mathrm{t,PFM}}(\bm{\theta}_t, \hat{\mathbf{v}}_t, \eta), { \scriptstyle\frac{\eta^2}{|\hat{\mathbf{v}}_t^{\top} \nabla^2 f(\bm{\theta}_t) \hat{\mathbf{v}}_t|}} \mathbf{I}\right),$$
where 
$$\bm{\mu}_{\mathrm{t,PFM}}(\bm{\theta}_t, \hat{\mathbf{v}}_t, \eta) = \bm{\theta}_t + \frac{\eta^2 \left(\nabla f( \bm{\theta}_t) \cdot \hat{\mathbf{v}}_t\right)\hat{\mathbf{v}}_t}{2 |\hat{\mathbf{v}}_t^{\top} \nabla^2 f(\bm{\theta}_t) \hat{\mathbf{v}}_t|}.$$
The equivalent conditional distribution for the reverse proposal, $q(\bm{\theta}_t|\bm{\theta}_*, \hat{\mathbf{v}}_*)$, requires sampling $\hat{\mathbf{v}}_*$ and swapping $\bm{\theta}_t$ for $\bm{\theta}_*$ and $\hat{\mathbf{v}}_t$ for $\hat{\mathbf{v}}_*$. Finally the MH acceptance ratio is evaluated, where the probabilities over the tangent vectors cancel each other due to their equal values, $q(\hat{\mathbf{v}}_*)=q(\hat{\mathbf{v}}_t)$. Figure \ref{fig:precon_fmala_example} illustrates the PC-FMALA proposal step. Compared to FMALA, we observe how the 2D covariances for $q(\bm{\theta}_*|\bm{\theta}_t, \hat{\mathbf{v}}_t)$ and $q(\bm{\theta}_t|\bm{\theta}_*, \hat{\mathbf{v}}_*)$ are now different sizes.

\subsubsection{Pre-conditioned line forward-mode MALA}
\begin{figure}[h!]
  \centering
  \begin{minipage}[t]{0.49\textwidth}
    \vspace{0pt} % Align top
    \centering
    \includegraphics[width=1.07\linewidth]{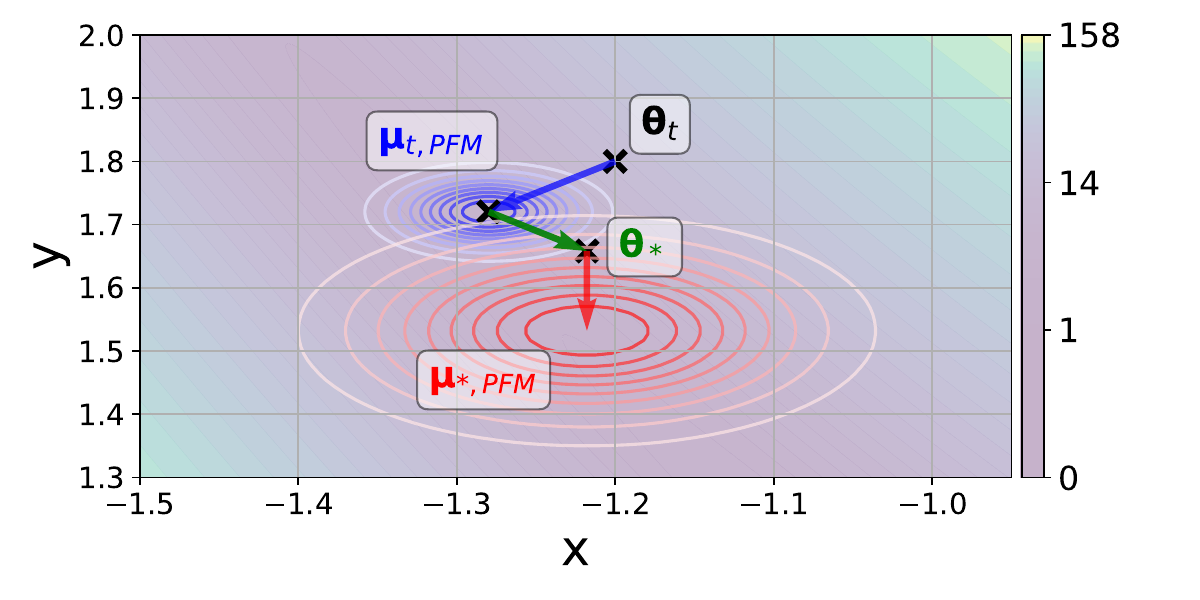}
    \caption{PC-FMALA single update step.} 
    % \textcolor{red}{Notice the change in size of the Gaussian proposals.}}
    \label{fig:precon_fmala_example}
  \end{minipage}
  \hfill
  \begin{minipage}[t]{0.49\textwidth}
    \vspace{0pt} % Align top
    \centering
    \includegraphics[width=1.07\linewidth]{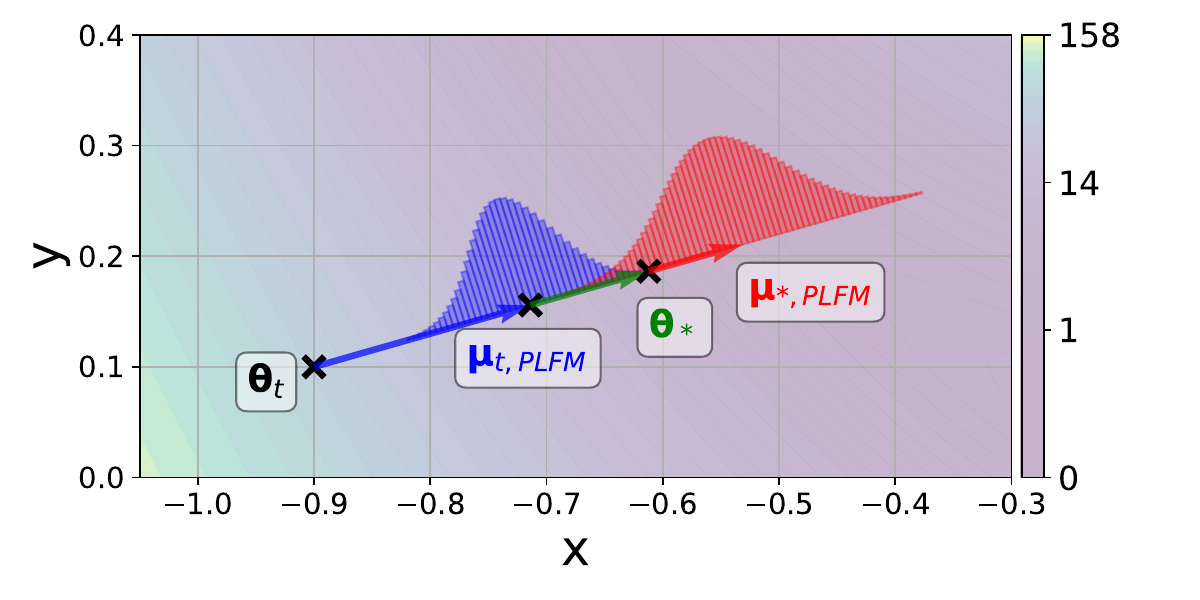}
    \caption{PC-Line-FMALA single update step.} %\textcolor{red}{Notice the change in size of the Gaussian proposals.}}
    \label{fig:precon_line_fmala_example}
  \end{minipage}
\end{figure}

While we might expect improved sampling efficiency from PC-FMALA, such as a higher acceptance rate and faster mixing, one potential drawback is that the Hessian information for the forward and reverse proposals are not evaluated along the same direction. This misalignment means that the benefits of incorporating position-specific geometry could be lessened in scenarios where the curvature varies significantly with the direction. Therefore we expect that constraining the direction using our line sampler could alleviate this potential issue. As a result, we introduce the pre-conditioned line forward-mode MALA (PC-L-FMALA). Just like for Line-FMALA, the sampler consists of two stages. The first stage samples the tangent vector as before, but the second stage now uses a proposal mechanism that relies on second-order information aligned with the tangent vector:
\begin{align}\label{eq:PreCon-Line-FMALA}
	\bm{\theta}_* = \bm{\theta} + \bigg( \frac{\eta^2 \nabla f( \bm{\theta}_t) \cdot \hat{\mathbf{v}}}{2 |\hat{\mathbf{v}}^{\top} \nabla^2 f(\bm{\theta}_t) \hat{\mathbf{v}}|}  + \frac{\eta z_t}{\sqrt{|\hat{\mathbf{v}}^{\top} \nabla^2 f(\bm{\theta}_t) \hat{\mathbf{v}}|}}\bigg)\hat{\mathbf{v}}.
\end{align}
Once again, we simplify the proposal mechanism to be scalar. Using the same notation as for Line-MALA ($\alpha_* = \bm{\theta}_* \cdot \hat{\mathbf{v}}$ and $\alpha_t = \bm{\theta}_t \cdot \hat{\mathbf{v}}$), we write Equation \eqref{eq:PreCon-Line-FMALA} as
\begin{equation}\label{eq:precon_line_FMALA_reduced}
	\alpha_* = \alpha_t + \frac{\eta^2 \nabla f( \bm{\theta}_t) \cdot \hat{\mathbf{v}}}{2 |\hat{\mathbf{v}}^{\top} \nabla^2 f(\bm{\theta}_t) \hat{\mathbf{v}}|}  + \frac{\eta z_t}{\sqrt{|\hat{\mathbf{v}}^{\top} \nabla^2 f(\bm{\theta}_t) \hat{\mathbf{v}}|}}.
\end{equation}
The proposal distributions are 
$q(\alpha_*|\alpha_t, \hat{\mathbf{v}}) = \mathcal{N}(\alpha_*; \mu_{\mathrm{t,PLFM}}(\alpha_t, \hat{\mathbf{v}}, \eta), \eta^2 /|\hat{\mathbf{v}}^{\top} \nabla^2 f(\bm{\theta}_t) \hat{\mathbf{v}}|)$, where $$\mu_{\mathrm{t,PLFM}}(\alpha_t, \hat{\mathbf{v}}, \eta) = \alpha_t + \frac{\eta^2 \nabla f( \bm{\theta}_t) \cdot \hat{\mathbf{v}}}{2 |\hat{\mathbf{v}}^{\top} \nabla^2 f(\bm{\theta}_t) \hat{\mathbf{v}}|},$$ 
and $q(\alpha_t|\alpha_*, \hat{\mathbf{v}})$, where $\alpha_t$ and $\bm{\theta}_t$ are swapped for $\alpha_*$ and $\bm{\theta}_*$. Figure \ref{fig:precon_line_fmala_example} illustrates the PC-Line-FMALA proposal mechanism in the same Rosenbrock function. The proposal distributions are 1D Gaussians over the line. Their standard deviations are now controlled by the second-order forward-mode quadratic Hessian term, such that the two proposals are of different shapes compared to Line-FMALA. 

\subsection{A note on bias and variance of the gradient estimator}
\citet{baydin2022gradients} proposed using tangent vectors drawn from a Normal distribution, resulting in the unbiased estimator $\mathbf{g}(\bm{\theta}) = (\nabla f( \bm{\theta}) \cdot \mathbf{v}) \mathbf{v}$ with expectation $\mathbb{E}[\mathbf{g}(\bm{\theta})] = \nabla f( \bm{\theta})$.
In our work, we instead sample tangent vectors uniformly from the unit sphere, $\hat{\mathbf{v}} \sim \mathrm{Uniform}(S^{D-1})$, which leads to cancellation in the MH acceptance step. As shown in App.~\ref{app:FGD_var}, the resulting gradient estimator $\hat{\mathbf{g}}(\bm{\theta})=(\nabla f( \bm{\theta}) \cdot \hat{\mathbf{v}}) \hat{\mathbf{v}}$ has the following elementwise expectation and variance:
\begin{equation}
    \mathbb{E}[\hat{g}_i(\bm{\theta})] = \frac{1}{D}\frac{\partial f}{\partial \theta_i}, \quad \mathrm{Var}\left(\hat{g}_i(\bm{\theta})\right) = \frac{1}{D(D+2)} \left( 2\frac{(D-1)}{D}\left[\frac{\partial f}{\partial \theta_i}\right]^2 + \sum_{j\neq i} \left[\frac{\partial f}{\partial \theta_j}\right]^2\right),
\end{equation}
where the expectation is biased by a multiplicative factor of $1/D$, and the variance contains additional $D$-dependent terms.
We can remove the bias by scaling $\hat{g}_i(\bm{\theta})$ by $D$, or equivalently, using tangent vectors $\sqrt{D}\hat{\mathbf{v}}$ and replacing all instances of $\hat{\mathbf{v}}$ in the above equations with $\sqrt{D}\hat{\mathbf{v}}$. This correction restores the unbiased expectation as in \citet{baydin2022gradients}, and asymptotically recovers the same variance as $D \to \infty$.
When applying this correction to our samplers, both line-based schemes remain unchanged. In FMALA, the JVP is multiplied by $D$, while in PreCon-FMALA, the proposal’s standard deviation is scaled by $1/\sqrt{D}$. In practice, for FMALA, we use a step size $\tilde{\eta}=\eta\sqrt{D}$ as it achieves superior performance. Full implementation details and derivations for these corrections are provided in App.~\ref{app:imp}.

\section{Experiments}\label{sec:exp}

\begin{table}[t!]
\small
    \centering
    \caption{Funnel distribution results across dimensions. 5 Chains, each of 10{,}000, 10 random seeds. Selected based on best KL performance.}\label{tab:funnel_data}
    \begin{tabular}{llrrrr}
      \toprule % from booktabs package
      &\bfseries Algorithm & \bfseries $\mathrm{D}_{\mathrm{KL}}(p(w)||q(w))$ $(\downarrow)$ & \bfseries $\mathrm{ESS}_w$ $(\uparrow)$ & \bfseries $\frac{1}{D} \sum_i\mathrm{ESS}_{\theta_i}$ $(\uparrow)$ & $\mathrm{KL}\ p$-value\\
      & & & & & (from MALA)\\
      \midrule % from booktabs package
      10D
      % \midrule % from booktabs package
      & FMALA & $0.202 \pm 0.125$ & $15.8 \pm 9.4$ & $140.5 \pm 77.2$ & 0.73\\
      & Line-FMALA & $0.176 \pm 0.089$ & $12.4 \pm 3.4$ & $107.9 \pm 50.3$  & 0.49 \\
      & PC-FMALA & $\mathbf{0.063 \pm 0.048}$ & $9.9 \pm 3.7$ & $132.9 \pm 78.3$ & 0.01 \\
      & PC-L-FMALA & $\mathbf{0.039 \pm 0.025}$ & $12.0 \pm 2.4$ & $366.5 \pm 87.4$ & 0.01
\\
      % \midrule
      & MALA (backprop.) & $0.160 \pm 0.075$ & $20.9 \pm 8.5$ & $111.2 \pm 45.6$ & - \\
      % & PreCon-MALA & - & - \\
      \midrule % from booktabs package
      
      50D
      % \midrule % from booktabs package
      & FMALA & $1.705 \pm 0.800$ & $7.2 \pm 1.3$ & $99.0 \pm 26.6$ & 0.00\\
      & Line-FMALA & $1.978 \pm 0.735$ & $7.9 \pm 1.0$ & $101.1 \pm 34.0$ & 0.00\\
      & PC-FMALA & $1.834 \pm 0.750$ & $6.9 \pm 0.7$ & $119.4 \pm 30.0$& 0.00\\
      & PC-L-FMALA & $\mathbf{0.917 \pm 0.584}$ & $7.0 \pm 1.1$ & $263.2 \pm 48.8$ & $0.79$ \\
      % \midrule
      & MALA (backprop.)& $\mathbf{0.549 \pm 0.327}$ & $10.7 \pm 2.2$ & $170.8 \pm 81.6$ & - \\
      % & PreCon-MALA & - & - \\
      \midrule % from booktabs package
      100D
      % \midrule % from booktabs package
      & FMALA & $7.150 \pm 2.627$ & $6.8 \pm 0.8$ & $110.6 \pm 15.9$& $0.00$\\
      & Line-FMALA & $5.929 \pm 1.913$ & $6.8 \pm 0.7$ & $99.6 \pm 18.0$& $0.00$\\
      & PC-FMALA & $7.877 \pm 2.990$ & $7.4 \pm 0.8$ & $83.2 \pm 9.50$& $0.00$ \\
      & PC-L-FMALA & $3.444 \pm 1.567$ & $6.6 \pm 0.4$ & $168.4 \pm 20.6$& $0.00$\\
      % \midrule
      & MALA (backprop.)& $\mathbf{1.366 \pm 0.531}$ & $9.4 \pm 1.4$ & $275.8 \pm 109.5$ & - \\
      % & PreCon-MALA & - & - \\
      \bottomrule % from booktabs package
    \end{tabular}
\end{table}

In the following experiments we investigate how the different variants of FMALA perform across multiple probabilistic models. 
Throughout this section, we compare to the original MALA which relies on backpropagation. 
We highlight that:
(1) reverse-mode AD is computationally more expensive than the forward-mode AD, which we see manifesting in the results when comparing MALA to our proposed forward-mode sampling algorithms, and (2) The forward-mode-only samplers can achieve comparable performance with MALA, and can even outperform MALA depeding on the structure of the probabilistic model. This combination of potential lower computational cost and comparable performance with MALA is highlighted throughout the experiments section via displaying MALA's increased wall-clock time and MALA's increased memory requirements, while also reporting on relevant metrics of sampler performance such as posterior predictive log-likelihood and accuracy.   

\subsection{Funnel distribution}\label{sec:funnel}
We evaluate sampler performance on Neal's funnel distribution \citep{neal2003slice}, $\pi (\bm{\theta}, w) = \prod_{i=1}^D \mathcal{N}(\theta_i \mid 0, \exp(-w)) \cdot \mathcal{N}(w \mid 0, 3^2)$, a well-known benchmark for hierarchical and ill-conditioned targets. To assess sample quality, we compare the KL divergence between the empirical marginal over $w$ and its known analytical form. Full setup and evaluation details, including statistical tests and ESS computation, are provided in App.~\ref{app:fun}.

\textbf{KL Performance.} For the $10D$ funnel, both PC-FMALA and PC-L-FMALA significantly outperform MALA in KL divergence ($p=0.01$). At $50D$, the performance of forward-mode samplers degrades relative to reverse-mode MALA, but PC-L-FMALA still performs on par with MALA. In $100D$, MALA achieves the best KL performance overall, though line-based variants outperform the other FMALA variants.

\begin{figure}[h!]\vspace{-5pt}
    \centering
    \includegraphics[width=.99\linewidth]{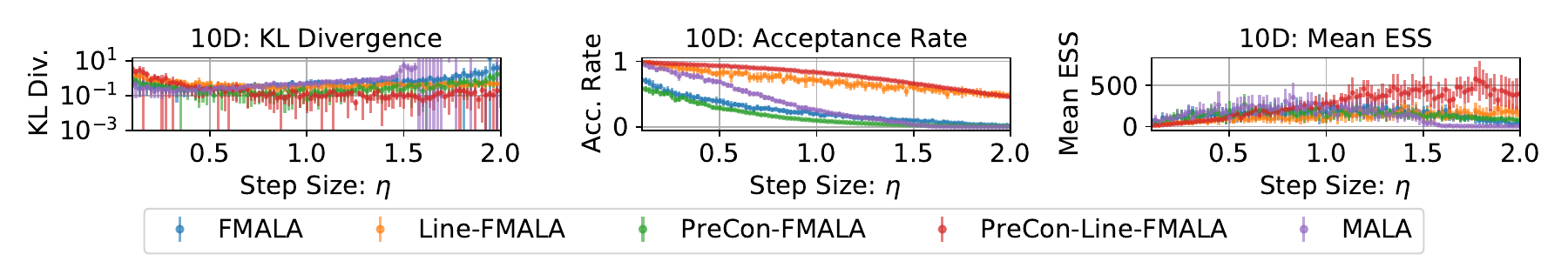}
    \caption{Hyperparameter Sensitivity. Error bars correspond to standard deviation across the 10 seeds at each step size.}
    \label{fig:10D_funnel}\vspace{-5pt}
\end{figure}

\textbf{Step Size Sensitivity.} Across dimensions, line-based samplers exhibit greater robustness to the step size $\eta$, as shown in Figure~\ref{fig:10D_funnel}. This insensitivity leads to higher acceptance rates across a wider range of $\eta$, simplifying tuning. PC-L-FMALA further increases the observed robustness to step size, compared to without pre-conditioning.

\subsection{Multinomial logistic regression}
We now introduce a multinomial logistic regression model with a Gaussian prior (see App. \ref{app:blg} for details). We use the MNIST dataset \citep{lecun1998gradient}, with $D_{\mathrm{in}}=784$ and $C=10$. We perform Bayesian optimization over the $\eta$ for each sampler, and run the sampler for $10^5$ samples. We burn the first $5\times10^4$ samples and then apply thinning by collecting $^1/_{100}$ samples thereafter. We calculate expected calibration error (ECE) using \texttt{torchmetrics.CalibrationError} with $100$ bins. 
 
 \textbf{Performance.} Table \ref{tab:mnist} displays the final performance over the test set, where all metrics use the posterior predictive mean. All the forward-mode samplers (except PC-FMALA) are competitive with the original MALA. This result highlights the significant potential of using the cheaper forward-mode samplers instead of samplers that use backpropagation. We did not see the same relative drop-off in performance compared to MALA that we saw in Sec. \ref{sec:funnel}. We hypothesize that certain geometries may not cause the forward-mode samplers to degrade in performance for higher dimension. We also highlight that all forward-mode samplers achieve a lower wall-clock time compared to MALA for average time per step. For example Line-FMALA achieves a 25~\% reduction in time.  

\begin{table}[h!]
\small
    \centering
    \caption{Multinomial Logistic Regression. The performance of all forward-mode samplers (except PC-MALA) is directly comparable with the reverse-mode sampler, MALA. The reduction in wall-clock time of the cheaper forward-mode samplers points to the potential of achieving comparable performance to MALA with less computation.}\label{tab:mnist}
    \begin{tabular}{lrrrr}
      \toprule % from booktabs package
      \bfseries Algorithm  
       & \bfseries NLL $(\downarrow)$ & \bfseries Acc. $(\uparrow)$& \bfseries  ECE $(\downarrow)$ & \bfseries Time/Step (ms) $(\downarrow)$\\
      \midrule % from booktabs package
      FMALA & 0.285 & 0.920 & 0.0187 &  69.2 $\pm$ 0.4 \\
      Line-FMALA & 0.290 & 0.919 & 0.0188 & 63.9 $\pm$ 0.8\\
      PC-FMALA & 0.363 & 0.891 & 0.0186 &  71.3 $\pm$ 1.0  \\
      PC-L-FMALA & 0.281 & 0.923 & 0.0175 & 64.8 $\pm$ 0.7 \\
      % \midrule
      MALA & 0.277 & 0.923 & 0.0189 & 85.7 $\pm$ 0.7 \\
      % PreCon-MALA & - & - \\
      \bottomrule % from booktabs package
    \end{tabular}
\end{table}

\textbf{Comment on PC-MALA.} While FMALA, Line-FMALA, and PC-Line-MALA perform well, PC-MALA underperforms the other samplers. This is not surprising given that PC-MALA estimates the Hessian in a randomly sampled direction for each of the forward and reverse proposals. In regions of variable curvature, where the value of $\hat{\mathbf{v}}_t^{\top} \nabla^2 f(\bm{\theta}_t) \hat{\mathbf{v}}_t$ depends closely on the update direction, the covariance may not be well matched to the target distribution. 

\subsection{Bayesian neural networks: regression}
We evaluate scalability to high-dimensional models with a Bayesian neural network (BNN), where $\tilde{\mathbf{Y}}=\mathrm{nn}(\mathbf{X};\bm{\theta})$ is a 5-layer fully connected network with ReLU activations and $D=40{,}701$ parameters. We use a Gaussian prior and likelihood with $\sigma_{\mathrm{prior}}=0.1$ and $\sigma_{\mathrm{lik}}=0.025$. The regression task uses a 400-point synthetic dataset from \citet{izmailov2020subspace}.

\begin{figure}[h!]
  \centering
  \begin{minipage}[t]{0.48\textwidth}
    \vspace{0pt} % force top alignment
    \small
    \centering
    \captionof{table}{BNN Regression. Both line forward-mode samplers outperform the other MCMC approaches. This further indicates the scalability of these forward-mode sampling schemes compared to reverse-mode. The Time/Step is evaluated across 100 steps with 5 repeats.}
    \label{tab:regression}
    \begin{tabular}{lrrr}
      \toprule
      \bfseries Algorithm & \bfseries NLL & \bfseries MSE & \bfseries Time/Step  \\
      & & & \bfseries (ms) \\
      \midrule
      FMALA & -0.168 & 0.011 & 81.8 $\pm$ 1.8 \\
      Line-FMALA & -0.671 & 0.011 & 35.4 $\pm$ 0.7 \\
      PC-FMALA & 12.371 & 0.022 & 88.4 $\pm$ 1.3 \\
      PC-L-FMALA & -0.691 & 0.011 & 38.1 $\pm$ 1.0 \\
      MALA & -0.458 & 0.011 & 37.0 $\pm$ 0.8 \\
      \bottomrule
    \end{tabular}
  \end{minipage}
  \hfill
  \begin{minipage}[t]{0.48\textwidth}
    \vspace{0pt} % force top alignment
    \centering
    \includegraphics[width=\linewidth]{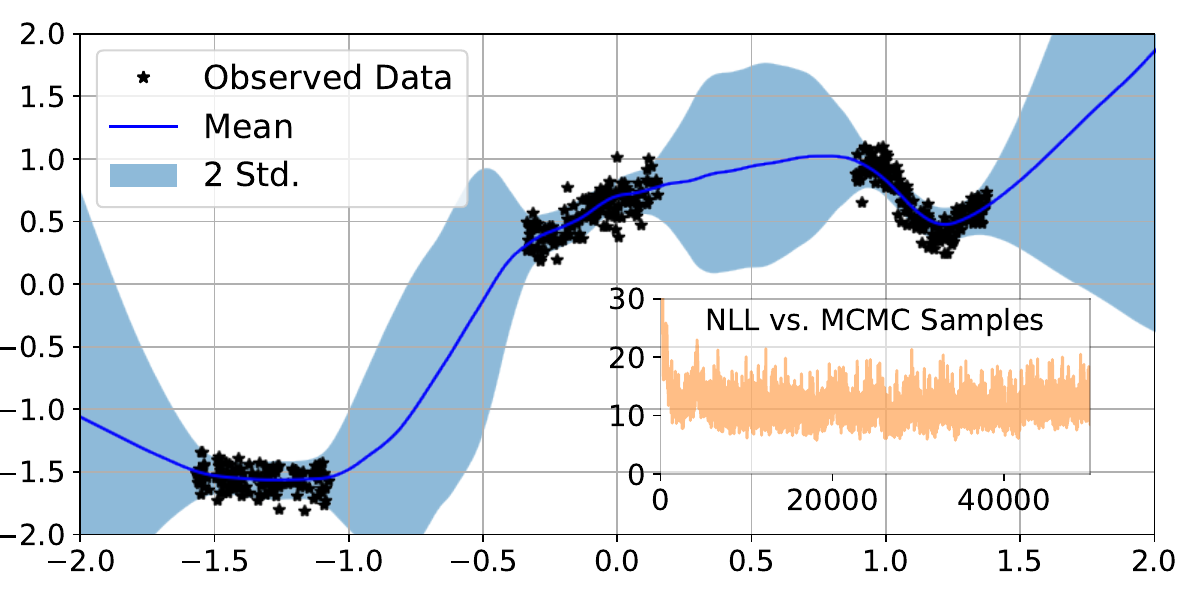}
    \caption{PreCon-Line-FMALA applied to a five-layer fully connected BNN. For the same number of samples ($5\times10^{4}$), achieves lowest MSE and NLL.}
    \label{fig:regression_pc_line_fmala}
  \end{minipage}
\end{figure}

All samplers are run for $5\times10^4$ iterations, with burn-in and thinning applied (see App. \ref{app:bnn_regression} for experimental details). Table~\ref{tab:regression} reports ensemble MSE and NLL on the training set. Forward-mode samplers (except PC-FMALA) scale well, with both line-based variants outperforming the others. Figure~\ref{fig:regression_pc_line_fmala} shows the qualitative performance of PreCon-Line-FMALA, including sample-based NLL and predictive uncertainty. The method yields well-calibrated uncertainty and rapid mixing.

\subsection{Bayesian CNN}\label{sec:cnn}

\begin{figure}[h!]
  \centering
  \begin{minipage}[t]{0.48\textwidth}
    \vspace{0pt} % force top alignment
    \small
    \centering
    \captionof{table}{$N=10{,}000$ results for CNN.}
    \label{tab:cifar10_accuracy_ece_short}
    \begin{tabular}{lcc}
        \toprule
        \textbf{Algorithm} & \textbf{Accuracy} & \textbf{ECE} \\
        \midrule
        FMALA         & 0.798 $\pm$ 0.002 & 0.025 $\pm$ 0.001 \\
        L-FMALA       & 0.797 $\pm$ 0.002 & 0.029 $\pm$ 0.003 \\
        PC-L-FMALA    & 0.796 $\pm$ 0.002 & 0.029 $\pm$ 0.002 \\
        MALA          & 0.798 $\pm$ 0.002 & 0.026 $\pm$ 0.003 \\
        % \midrule
        % Base: SGD           & 0.794 & 0.036 \\
        \bottomrule
    \end{tabular}

    \centering
    \captionof{table}{Time/step for two training data sizes. Note MALA memory error for $N=30{,}000$.}
    \label{tab:step_time_combined_short}
    \begin{tabular}{lcc}
        \toprule
        \textbf{Algorithm} & \bfseries {$N=10{,}000$} & \bfseries {$N=30{,}000$} \\
        & \textbf{Time/Step (s)} & \textbf{Time/Step (s)} \\
        \midrule
        FMALA            & 0.40 $\pm$ 0.00 & 0.96 $\pm$ 0.00 \\
        Line-FMALA       & 0.34 $\pm$ 0.00 & 0.91 $\pm$ 0.00 \\
        % PC-FMALA         & 0.53 $\pm$ 0.00 & 1.35 $\pm$ 0.00 \\
        PC-L-FMALA       & 0.47 $\pm$ 0.00 & 1.29 $\pm$ 0.00 \\
        MALA             & 0.52 $\pm$ 0.00 & \textbf{Mem. Error}  \\
        \bottomrule
    \end{tabular}

  \end{minipage}
  \hfill
  \begin{minipage}[t]{0.48\textwidth}
    \vspace{0pt} % force top alignment
    \centering
  % \vspace{-16pt} % Adjust if needed to align better with text
  \includegraphics[width=\linewidth]{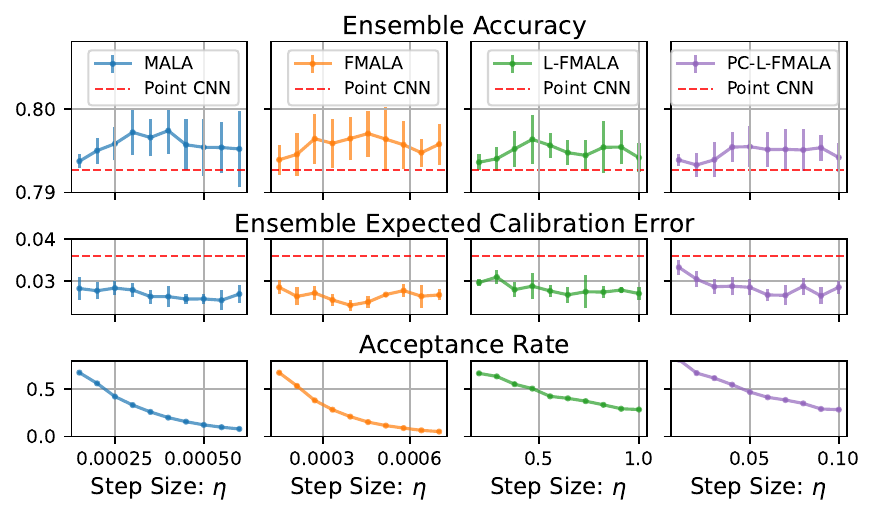}
  \caption{CNN classification for CIFAR10. We initialize from a $2{,}396{,}330$ parameter CNN and observe improved accuracy and ECE performance. This experiment highlights the scalability of forward-mode AD to larger $D$.}
  \label{fig:cifar_pretrain}
  % \vspace{-10pt} % Optional tweak to reduce vertical gap after
  \end{minipage}
\end{figure}

We now scale to a CNN with $2{,}396{,}330$ parameters trained on $10{,}000$ CIFAR-10 images \citep{krizhevsky2009learning}, initialized via stochastic gradient descent (SGD).\footnote{Limited to $10{,}000$ images due to GPU memory constraints.} We use $\sigma_{\mathrm{prior}}=10$, consistent with the SGD weight decay, and a cross-entropy likelihood.
Figure~\ref{fig:cifar_pretrain} shows a step size grid search for ensemble accuracy and ECE (1,000 samples, averaged over 5 seeds), demonstrating the scalability of forward-mode MALA. All samplers improve accuracy and calibration at optimal $\eta$ (Table \ref{tab:cifar10_accuracy_ece_short}), with forward-mode variants offering clear compute advantages (Table \ref{tab:step_time_combined_short}).
Additional wall-clock times and memory comparisons are provided in App.~\ref{sec:cnn}, where we show \textbf{MALA fails with a RESOURCE EXHAUSTED (memory) error for greater than $10{,}000$ images}, while forward-mode samplers run successfully. This highlights the memory advantage of forward-mode compared to reverse-mode AD approaches.

\section{Limitations}
Our proposed forward-mode samplers show strong potential to reduce the computational cost of gradient-based sampling, however there are also important trade-offs that we have observed. In particular, we see a strong dependence on structure of the target distribution. For the funnel distribution, forward-mode samplers underperform at higher dimensions, though they match or exceed performance for smaller $D$. For multinomial logistic regression, the sample quality is comparable in terms of the performance metrics, but our best forward-mode sampler achieves a 25~\% reduction in wall-clock time. For the BNN regression experiment, the line-based forward samplers outperform MALA in NLL, though with less wall-clock advantage, likely due to the smaller dataset size. Most notably, in our highest-dimensional model, Line-FMALA yields a 34~\% time reduction, while all forward-mode samplers avoid the memory error encountered by MALA for $N\geq30{,}000$, despite MALA having marginally better accuracy when $N=10{,}000$ (Tables \ref{tab:cifar10_accuracy_ece_short} and \ref{tab:step_time_combined_short}). These results suggest the relative benefit of forward-mode methods depends on the geometry and memory demands of the target distribution, underscoring their promise in high-dimensional or memory-constrained settings.   

\section{Conclusion}\label{sec:conc}
Overall, this paper is the first to introduce the benefits of using forward-mode AD in MCMC sampling schemes.
We have shown that forward-mode MALA is competitive with reverse-mode MALA, and can even outperform MALA depending on the probabilistic model.
This finding is significant since the runtime cost and memory cost of forward-mode approaches are substantially lower than their reverse-mode counterparts. To connect forward-mode AD with MALA, we defined four new sampling schemes, including a novel two-stage line-based sampling scheme and the use of second-order forward-mode AD to precondition FMALA. We found that PreCon-Line-MALA, which combines both algorithmic contributions, is often the most performant of the forward-mode samplers according to the task-specific metrics across the large variation in models.

% \begin{contributions} % will be removed in pdf for initial submission 
% 					  % (without ‘accepted’ option in \documentclass)
%                       % so you can already fill it to test with the
%                       % ‘accepted’ class option
%     Briefly list author contributions. 
%     This is a nice way of making clear who did what and to give proper credit.
%     This section is optional.

%     H.~Q.~Bovik conceived the idea and wrote the paper.
%     Coauthor One created the code.
%     Coauthor Two created the figures.
% \end{contributions}

\begin{ack}
We would like to acknowledge the valuable discussions, feedback, and resources provided by our colleagues and external collaborators throughout the process. This material is based upon work supported by the United States Air Force and DARPA under Contract No. FA8750-23-C-0519 and HR0011-24-9-0424, and the U.S. Army Research Laboratory under Cooperative Research Agreement W911NF-17-2-0196. Any opinions, findings and conclusions or recommendations expressed in this material are those of the author(s) and do not necessarily reflect the views of the United States Air Force, DARPA, the U.S. Army Research Laboratory, or the United States Government.
\end{ack}

% References
\bibliography{neurips}

%%%%%%%%%%%%%%%%%%%%%%%%%%%%%%%%%%%%%%%%%%%%%%%%%%%%%%%%%%%%
\newpage
\appendix
\bibliographystyle{plainnat}

\section{Algorithms}\label{app:alg}
\begin{algorithm}
\caption{Forward-Mode Metropolis Adjusted Langevin Dynamics (FMALA)}
\label{alg:fmala}
\begin{algorithmic}
\REQUIRE Target distribution $f(\bm{\theta}) = \log \pi(\bm{\theta})$, step size $\eta$, initial state $\bm{\theta}_0$
\FOR{$t = 0, 1, 2, \dots$ until convergence}
    \STATE $\hat{\mathbf{v}}_t \sim \mathrm{Uniform}(S^{D-1})$
    \STATE // Evaluate forward-mode step at $\bm{\theta}_t$
    \STATE $f(\bm{\theta}_t), \quad \nabla f(\bm{\theta}_t) \cdot \hat{\mathbf{v}}_t \leftarrow F_1(\bm{\theta}_t, \hat{\mathbf{v}}_t)$
    \STATE // Generate proposal using Langevin dynamics
    \STATE $\bm{\mu}_{\mathrm{FM}}(\bm{\theta}_t, \hat{\mathbf{v}}_t, \eta) = \bm{\theta}_t + \frac{\eta^2}{2} \left(\nabla f( \bm{\theta}_t) \cdot \hat{\mathbf{v}}_t\right)\hat{\mathbf{v}}_t$
    \STATE $\bm{\theta}_* = \bm{\mu}_{\mathrm{FM}}(\bm{\theta}_t, \hat{\mathbf{v}}_t, \eta) + \eta \mathbf{z}_t$, where $\mathbf{z}_t \sim \mathcal{N}(\mathbf{0}, \mathbf{I})$
    \STATE // Evaluate components of $q(\bm{\theta}_t|\bm{\theta}_*, \hat{\mathbf{v}}_*)$
    \STATE $\hat{\mathbf{v}}_* \sim \mathrm{Uniform}(S^{D-1})$
    \STATE $f(\bm{\theta}_*), \quad \nabla f(\bm{\theta}_*) \cdot \hat{\mathbf{v}}_* \leftarrow F_1(\bm{\theta}_*, \hat{\mathbf{v}}_*)$
    \STATE $\bm{\mu}_{\mathrm{FM}}(\bm{\theta}_*, \hat{\mathbf{v}}_*, \eta) = \bm{\theta}_* + \frac{\eta^2}{2} \left(\nabla f( \bm{\theta}_*) \cdot \hat{\mathbf{v}}_*\right)\hat{\mathbf{v}}_*$
    \STATE // Compute Metropolis acceptance probability
    \STATE $\gamma = \min\left(0, \left[f(\bm{\theta}_*) + \log \mathcal{N}\left(\bm{\theta}_t; \bm{\mu}_{\mathrm{FM}}(\bm{\theta}_*, \hat{\mathbf{v}}_*, \eta) , \eta^2 \mathbf{I}\right)\right] \right.
    % $
    % \STATE $\quad \quad \quad \quad \quad \quad \quad \quad 
    - \left.\left[f(\bm{\theta}_t) + \log \mathcal{N}\left(\bm{\theta}_*; \bm{\mu}_{\mathrm{FM}}(\bm{\theta}_t, \hat{\mathbf{v}}_t, \eta) , \eta^2 \mathbf{I} \right) \right]\right)$    
    \STATE // Accept or reject the proposal
    \STATE $u \sim \text{Uniform}(0, 1)$
    \IF{$\log u < \gamma$}
        \STATE $\bm{\theta}_{t+1} = \bm{\theta}_*$
    \ELSE
        \STATE $\bm{\theta}_{t+1} = \bm{\theta}_t$
    \ENDIF
\ENDFOR
\end{algorithmic}
\end{algorithm}

%%%%%%%%%%%%%%%%%%%%%% Line FMALA ###############################

\begin{algorithm}
\caption{Line Forward-Mode Metropolis Adjusted Langevin Dynamics (Line-FMALA)}
\label{alg:line_fmala}
\begin{algorithmic}
\REQUIRE Target distribution $f(\bm{\theta}) = \log \pi(\bm{\theta})$, step size $\eta$, initial state $\bm{\theta}_0$
\FOR{$t = 0, 1, 2, \dots$ until convergence}
    \STATE // Stage 1: Sample tangent direction
    \STATE $\hat{\mathbf{v}} \sim \mathrm{Uniform}(S^{D-1})$
    \STATE // Stage 2: Metropolis-Hastings along sampled direction
    \STATE // Evaluate forward-mode step at $\bm{\theta}_t$
    \STATE $f(\bm{\theta}_t), \quad \nabla f(\bm{\theta}_t) \cdot \hat{\mathbf{v}} \leftarrow F_1(\bm{\theta}_t, \hat{\mathbf{v}})$
    \STATE // Generate proposal using Langevin dynamics
    \STATE $\mu_{\mathrm{LFM}}(\alpha_t, \hat{\mathbf{v}}, \eta) = \alpha_t + \frac{\eta^2}{2} \nabla f( \bm{\theta}_t) \cdot \hat{\mathbf{v}}$, where $\alpha_t = \bm{\theta}_t \cdot \hat{\mathbf{v}}$ as in Equation \eqref{eq:line_FMALA_reduced}.
    \STATE $\alpha_* = \mu_{\mathrm{LFM}}(\alpha_t, \hat{\mathbf{v}}, \eta) + \eta z_t$, where $z_t \sim \mathcal{N}(0, 1)$
    \STATE // Evaluate components of $q(\alpha_t|\alpha_*, \hat{\mathbf{v}})$
    \STATE $f(\bm{\theta}_*), \quad \nabla f(\bm{\theta}_*) \cdot \hat{\mathbf{v}} \leftarrow F_1(\bm{\theta}_*, \hat{\mathbf{v}})$
    \STATE $\mu_{\mathrm{LFM}}(\alpha_*, \hat{\mathbf{v}}, \eta) = \alpha_* + \frac{\eta^2}{2} \nabla f( \bm{\theta}_*) \cdot \hat{\mathbf{v}}$
    \STATE // Compute Metropolis acceptance probability
    \STATE $\gamma = \min\left(0, \left[f(\bm{\theta}_*) + \log \mathcal{N}\left(\alpha_t; \mu_{\mathrm{LFM}}(\alpha_*, \hat{\mathbf{v}}, \eta) , \eta^2\right)\right] \right.
    % $
    % \STATE $\quad \quad \quad \quad \quad \quad \quad \quad 
    - \left.\left[f(\bm{\theta}_t) + \log \mathcal{N}\left(\alpha_*; \mu_{\mathrm{LFM}}(\alpha_t, \hat{\mathbf{v}}, \eta) , \eta^2 \right) \right]\right)$    
    \STATE // Accept or reject the proposal
    \STATE $u \sim \text{Uniform}(0, 1)$
    \IF{$\log u < \gamma$}
        \STATE $\bm{\theta}_{t+1} = \bm{\theta}_*$
    \ELSE
        \STATE $\bm{\theta}_{t+1} = \bm{\theta}_t$
    \ENDIF
\ENDFOR
\end{algorithmic}
\end{algorithm}

%%%%%%%%%%%%%%%%%%%%%% PreCon- FMALA ###############################

\begin{algorithm}
\caption{Pre-conditioned Forward-Mode Metropolis Adjusted Langevin Dynamics (PC-FMALA)}
\label{alg:fmala}
\begin{algorithmic}
\REQUIRE Target distribution $f(\bm{\theta}) = \log \pi(\bm{\theta})$, step size $\eta$, initial state $\bm{\theta}_0$
\FOR{$t = 0, 1, 2, \dots$ until convergence}
    \STATE $\hat{\mathbf{v}}_t \sim \mathrm{Uniform}(S^{D-1})$
    \STATE //Evaluate second-order forward-mode step at $\bm{\theta}_t$
    \STATE $f(\bm{\theta}_t), \quad \nabla f(\bm{\theta}_t) \cdot \hat{\mathbf{v}}_t, \quad \hat{\mathbf{v}}_t^{\top} \nabla^2 f(\bm{\theta}_t)\hat{\mathbf{v}}_t \quad \leftarrow F_2(\bm{\theta}_t, \hat{\mathbf{v}}_t)$
    \STATE // Generate proposal using Langevin dynamics
    \STATE $\bm{\mu}_{\mathrm{PFM}}(\bm{\theta}_t, \hat{\mathbf{v}}_t, \eta) = \bm{\theta}_t + \frac{\eta^2 \left(\nabla f( \bm{\theta}_t) \cdot \hat{\mathbf{v}}_t\right)\hat{\mathbf{v}}_t}{2 |\hat{\mathbf{v}}_t^{\top} \nabla^2 f(\bm{\theta}_t) \hat{\mathbf{v}}_t|}$
    \STATE $\bm{\theta}_* = \bm{\mu}_{\mathrm{PFM}}(\bm{\theta}_t, \hat{\mathbf{v}}_t, \eta) + \eta \sqrt{\left(|\hat{\mathbf{v}}_t^{\top} \nabla^2 f(\bm{\theta}_t) \hat{\mathbf{v}}_t|\right)^{-1}}\mathbf{z}_t$
    \STATE // Evaluate components of $q(\bm{\theta}_t|\bm{\theta}_*, \hat{\mathbf{v}}_*)$
    \STATE $\hat{\mathbf{v}}_* \sim \mathrm{Uniform}(S^{D-1})$
    \STATE $f(\bm{\theta}_*), \quad \nabla f(\bm{\theta}_*) \cdot \hat{\mathbf{v}}_*, \quad \hat{\mathbf{v}}_*^{\top} \nabla^2 f(\bm{\theta}_*)\hat{\mathbf{v}}_* \quad \leftarrow F_2(\bm{\theta}_*, \hat{\mathbf{v}}_*)$
    \STATE $\bm{\mu}_{\mathrm{PFM}}(\bm{\theta}_*, \hat{\mathbf{v}}_*, \eta) = \bm{\theta}_* + \frac{\eta^2 \left(\nabla f( \bm{\theta}_*) \cdot \hat{\mathbf{v}}_*\right)\hat{\mathbf{v}}_*}{2 |\hat{\mathbf{v}}_*^{\top} \nabla^2 f(\bm{\theta}_*) \hat{\mathbf{v}}_*|}$
    \STATE // Compute Metropolis acceptance probability
    \STATE $\gamma = \min\left(0, \left[f(\bm{\theta}_*) + \log \mathcal{N}\left(\bm{\theta}_t; \bm{\mu}_{\mathrm{PFM}}(\bm{\theta}_*, \hat{\mathbf{v}}_*, \eta) , \frac{\eta^2}{\left(|\hat{\mathbf{v}}_*^{\top} \nabla^2 f(\bm{\theta}_*) \hat{\mathbf{v}}_*|\right)}\mathbf{I}\right)\right] \right.
    $
    \STATE $\quad \quad \quad \quad \quad \quad \quad \quad 
    - \left.\left[f(\bm{\theta}_t) + \log \mathcal{N}\left(\bm{\theta}_*; \bm{\mu}_{\mathrm{PFM}}(\bm{\theta}_t, \hat{\mathbf{v}}_t, \eta) , \frac{\eta^2}{\left(|\hat{\mathbf{v}}_t^{\top} \nabla^2 f(\bm{\theta}_t) \hat{\mathbf{v}}_t|\right)}\mathbf{I} \right) \right]\right)$    
    \STATE // Accept or reject the proposal
    \STATE $u \sim \text{Uniform}(0, 1)$
    \IF{$\log u < \gamma$}
        \STATE $\bm{\theta}_{t+1} = \bm{\theta}_*$
    \ELSE
        \STATE $\bm{\theta}_{t+1} = \bm{\theta}_t$
    \ENDIF
\ENDFOR
\end{algorithmic}
\end{algorithm}

%%%%%%%%%%%%%%%%%%%%%% PreCon-Line FMALA ###############################

\begin{algorithm}
\caption{Pre-conditioned Line Forward-Mode Metropolis Adjusted Langevin Dynamics (PC-Line-FMALA)}
\label{alg:precon_line_fmala}
\begin{algorithmic}
\REQUIRE Target distribution $f(\bm{\theta}) = \log \pi(\bm{\theta})$, step size $\eta$, initial state $\bm{\theta}_0$
\FOR{$t = 0, 1, 2, \dots$ until convergence}
    \STATE // Stage 1: Sample tangent direction
    \STATE $\hat{\mathbf{v}} \sim \mathrm{Uniform}(S^{D-1})$
    \STATE // Stage 2: Metropolis-Hastings along sampled direction
    \STATE // Evaluate second-order forward-mode step at $\bm{\theta}_t$
    \STATE $f(\bm{\theta}_t), \quad \nabla f(\bm{\theta}_t) \cdot \hat{\mathbf{v}}_t, \quad \hat{\mathbf{v}}_t^{\top} \nabla^2 f(\bm{\theta}_t)\hat{\mathbf{v}}_t \quad \leftarrow F_2(\bm{\theta}_t, \hat{\mathbf{v}}_t)$
    \STATE // Generate proposal using Langevin dynamics
    \STATE $\mu_{\mathrm{PLFM}}(\alpha_t, \hat{\mathbf{v}}, \eta) = \alpha_t + \frac{\eta^2 \nabla f( \bm{\theta}_t) \cdot \hat{\mathbf{v}}}{2 |\hat{\mathbf{v}}^{\top} \nabla^2 f(\bm{\theta}_t) \hat{\mathbf{v}}|}$, where $\alpha_t = \bm{\theta}_t \cdot \hat{\mathbf{v}}$ as in Equation \eqref{eq:precon_line_FMALA_reduced}.
    \STATE $\alpha_* = \mu_{\mathrm{PLFM}}(\alpha_t, \hat{\mathbf{v}}, \eta) + \frac{\eta z_t}{\sqrt{|\hat{\mathbf{v}}^{\top} \nabla^2 f(\bm{\theta}_t) \hat{\mathbf{v}}|}}$, where $z_t \sim \mathcal{N}(0, 1)$
    \STATE // Evaluate components of $q(\alpha_t|\alpha_*, \hat{\mathbf{v}})$
    \STATE $f(\bm{\theta}_*), \quad \nabla f(\bm{\theta}_*) \cdot \hat{\mathbf{v}}_*, \quad \hat{\mathbf{v}}_*^{\top} \nabla^2 f(\bm{\theta}_*)\hat{\mathbf{v}}_* \quad \leftarrow F_2(\bm{\theta}_*, \hat{\mathbf{v}}_*)$
    \STATE $\mu_{\mathrm{PLFM}}(\alpha_*, \hat{\mathbf{v}}, \eta) = \alpha_* + \frac{\eta^2 \nabla f( \bm{\theta}_*) \cdot \hat{\mathbf{v}}}{2 |\hat{\mathbf{v}}^{\top} \nabla^2 f(\bm{\theta}_*) \hat{\mathbf{v}}|}$
    \STATE // Compute Metropolis acceptance probability
    \STATE $\gamma = \min\left(0, \left[f(\bm{\theta}_*) + \log \mathcal{N}\left(\alpha_t; \mu_{\mathrm{PLFM}}(\alpha_*, \hat{\mathbf{v}}, \eta) , \frac{\eta^2}{|\hat{\mathbf{v}}^{\top} \nabla^2 f(\bm{\theta}_*) \hat{\mathbf{v}}|} \right)\right] \right.
    $
    \STATE $\quad \quad \quad \quad \quad \quad \quad \quad 
    - \left.\left[f(\bm{\theta}_t) + \log \mathcal{N}\left(\alpha_*; \mu_{\mathrm{PLFM}}(\alpha_t, \hat{\mathbf{v}}, \eta) , \frac{\eta^2}{|\hat{\mathbf{v}}^{\top} \nabla^2 f(\bm{\theta}_t) \hat{\mathbf{v}}|} \right) \right]\right)$    
    \STATE // Accept or reject the proposal
    \STATE $u \sim \text{Uniform}(0, 1)$
    \IF{$\log u < \gamma$}
        \STATE $\bm{\theta}_{t+1} = \bm{\theta}_*$
    \ELSE
        \STATE $\bm{\theta}_{t+1} = \bm{\theta}_t$
    \ENDIF
\ENDFOR
\end{algorithmic}
\end{algorithm}

\newpage

\section{Variance Analysis of Forward Gradient Estimator}\label{app:FGD_var}

If we start with the definition from \citet{baydin2022gradients} of the forward gradient:
$$
g_i(\bm{\theta}) = \frac{\partial f}{\partial \theta_i} v_i^2 + \sum_{j\neq i} \frac{\partial f}{\partial \theta_j} v_i v_j,
$$
Since FMALA and its variants use $\hat{\mathbf{v}} \sim \mathrm{Uniform}(S^{D-1})$, the expectation is now given by $\mathbb{E}\left[\hat{g}_i(\bm{\theta})\right] = \frac{1}{D}\frac{\partial f}{\partial \theta_i}$, where the first moment for $\hat{\mathbf{v}} \sim \mathcal{U}(S^{D-1})$ is $\mathbb{E}[\hat{v}_i] = 0$, and the second moment $\mathbb{E}[\hat{v}_i^2] = \frac{1}{D}$, (with cross terms $\mathbb{E}[\hat{v}_i^2] = 0$). $\mathbb{E}\left[\hat{g}_i(\bm{\theta})\right]$ is now a biased estimate of the gradient, where it is scaled by the inverse of the parameter dimension. This is accounted for within the FMALA algorithm through use of the learning rate $\eta$.

It is of interest as to how the variance of this estimate behaves, i.e. $\mathrm{Var}\left(g_i(\bm{\theta})\right)$. Using the definition, $\mathrm{Var}\left(X\right) = \mathbb{E}[X^2] - \mathbb{E}^2[X]$, we derive each component as:
$$
\mathbb{E}^2[\hat{g}_i(\bm{\theta})] = \frac{1}{D^2}\left[\frac{\partial f}{\partial \theta_i}\right]^2,
$$
and
\begin{align}
    \mathbb{E}[\hat{g}_i(\bm{\theta})^2] &= \mathbb{E}\left[\left[\frac{\partial f}{\partial \theta_i}\right]^2 \hat{v}_i^4 + 2 \frac{\partial f}{\partial \theta_i} \sum_{j\neq i} \frac{\partial f}{\partial \theta_j} \hat{v}_i ^3 \hat{v}_j + \left(\sum_{j\neq i} \frac{\partial f}{\partial \theta_j} \hat{v}_i \hat{v}_j\right)^2\right].\notag 
\end{align}
% https://chatgpt.com/share/e1c37429-5aed-43ed-96a5-69c2a519231b
We expand the last term in the expectation on the right-hand side as: 
$$
\left(\sum_{j\neq i} \frac{\partial f}{\partial \theta_j} \hat{v}_i \hat{v}_j\right)^2 = \hat{v}_i ^ 2 \left(\sum_{j\neq i} \left[\frac{\partial f}{\partial \theta_j}\right]^2 \hat{v}_j^2 + 2 \sum_{j\neq i} \sum_{k\neq i, k>j} \frac{\partial f}{\partial \theta_j} \frac{\partial f}{\partial \theta_k} \hat{v}_j \hat{v}_k \right),
$$
then moving the expectation inside the square brackets and using the identities, $\mathbb{E}[X^2] = \frac{1}{D}$, $\mathbb{E}[X^4] = \frac{3}{D(D+2)}$, $\mathbb{E}[X^3 Y] = 0$, $\mathbb{E}[X^2 Y^2] = \frac{1}{D(D+2)}$, and $\mathbb{E}[X^2 Y Z] = 0$  for the uniform sphere distribution gives:
\begin{align}
    \mathbb{E}[\hat{g}_i(\bm{\theta})^2] &= \left[\frac{\partial f}{\partial \theta_i}\right]^2 \left[\frac{3}{D(D+2)}\right] + 2 \frac{\partial f}{\partial \theta_i} \sum_{j\neq i} \frac{\partial f}{\partial \theta_j} [0] + \sum_{j\neq i} \left[\frac{\partial f}{\partial \theta_j}\right]^2 \left[\frac{1}{D(D+2)}\right] \notag \\&\quad\quad\quad\quad\quad\quad\quad\quad\quad+ 2 \sum_{j\neq i} \sum_{k\neq i, k>j} \frac{\partial f}{\partial \theta_j} \frac{\partial f}{\partial \theta_k} [0]\notag\\
    &= \frac{1}{D(D+2)} \left(3 \left[\frac{\partial f}{\partial \theta_i}\right]^2 + \sum_{j\neq i} \left[\frac{\partial f}{\partial \theta_j}\right]^2\right).\notag
\end{align}
Finally, the variance is given by
$$
\mathrm{Var}\left(\hat{g}_i(\bm{\theta})\right) = \frac{1}{D(D+2)} \left( 2\frac{(D-1)}{D}\left[\frac{\partial f}{\partial \theta_i}\right]^2 + \sum_{j\neq i} \left[\frac{\partial f}{\partial \theta_j}\right]^2\right).
$$
If we multiply the original estimator by $D$ to make it an unbiased estimate, $\mathbb{E}\left[D \cdot \hat{g}_i(\bm{\theta})\right] = \frac{\partial f}{\partial \theta_i}$, then $\mathrm{Var}\left(D \cdot \hat{g}_i(\bm{\theta})\right) = D^2\mathrm{Var}\left( \hat{g}_i(\bm{\theta})\right)$. Then as $D\rightarrow\infty$, 
$$
\lim_{D \to \infty}
\mathrm{Var}\left(D \cdot \hat{g}_i(\bm{\theta})\right) =  2\left[\frac{\partial f}{\partial \theta_i}\right]^2 + \sum_{j\neq i} \left[\frac{\partial f}{\partial \theta_j}\right]^2.
$$
This result shows that the variance of the unbiased estimator, $D \cdot \hat{g}_i(\bm{\theta})$, is equivalent to the variance of the original FGD estimator of \cite{baydin2022gradients}.

\subsection{Bias Correction}\label{app:imp}

If we multiply each tangent vector by $\sqrt{D}$, then we recover the unbiased estimate $g_i(\bm{\theta}) = D \cdot g_i(\bm{\theta}) = \left(\nabla f( \bm{\theta}_t) \cdot \sqrt{D} \hat{\mathbf{v}}_t\right) \sqrt{D}\hat{\mathbf{v}}_t$. We now derive the bias corrected proposal mechanism which is used as a drop in replacement for the proposed samplers. Note that Line-FMALA and PC-Line-FMALA are unaffected by this correction.

\subsubsection{FMALA}
The corrected proposal mechanism is given by
\begin{equation}
	\bm{\theta}_* = \bm{\theta}_t + \frac{(\eta\sqrt{D})^2}{2} \left(\nabla f( \bm{\theta}_t) \cdot \hat{\mathbf{v}}_t\right)\hat{\mathbf{v}}_t + \eta \mathbf{z}_t.
\end{equation}
To counter the potential imbalance between the gradient term and the noise term, we define $\tilde{\eta} :=\eta\sqrt{D}$ and couple the step size with the square-root of the dimension of the parameter space. This provided significant improvement in the experiments and is used throughout the paper.
% \textcolor{red}{Therefore we can just absorb $\sqrt{D}$ into $\eta$ without too many issues... Try coupling $\eta$ and $\sqrt{D}$ just to see. }

\subsubsection{Line-FMALA}
For Line-FMALA, the new proposal mechanism,
\begin{align}
	\bm{\theta}_* &= \bm{\theta}_t + \left(\frac{(\eta \sqrt{D})^2}{2} \nabla f( \bm{\theta}_t) \cdot \hat{\mathbf{v}} + \eta \sqrt{D} z_t\right)\hat{\mathbf{v}},\\
    &= \bm{\theta}_t + \left(\frac{\tilde{\eta}^2}{2} \nabla f( \bm{\theta}_t) \cdot \hat{\mathbf{v}} + \tilde{\eta} z_t\right)\hat{\mathbf{v}}, \quad \text{where } \tilde{\eta} := \eta \sqrt{D},
\end{align}
is equivalent to Eq. \eqref{eq:line_FMALA}, but with $\tilde{\eta} := \eta \sqrt{D}$.

% \textcolor{red}{Therefore, since the bias correction only appears as a multiple of the step size, we recover the original Line-FMALA equation, with the bias just absorbed into the step size.} 

\subsubsection{PC-FMALA}
For PC-FMALA, we get:
\begin{align}
	\bm{\theta}_* &= \bm{\theta}_t + \frac{D\eta^2}{2 D|\hat{\mathbf{v}}_t^{\top} \nabla^2 f(\bm{\theta}_t) \hat{\mathbf{v}}_t|} \left(\nabla f( \bm{\theta}_t) \cdot \hat{\mathbf{v}}_t\right)\hat{\mathbf{v}}_t + \eta \sqrt{\left(D|\hat{\mathbf{v}}_t^{\top} \nabla^2 f(\bm{\theta}_t) \hat{\mathbf{v}}_t|\right)^{-1}}\mathbf{z}_t \notag \\
    &= \bm{\theta}_t + \frac{\eta^2}{2|\hat{\mathbf{v}}_t^{\top} \nabla^2 f(\bm{\theta}_t) \hat{\mathbf{v}}_t|} \left(\nabla f( \bm{\theta}_t) \cdot \hat{\mathbf{v}}_t\right)\hat{\mathbf{v}}_t + \eta \sqrt{\left(D|\hat{\mathbf{v}}_t^{\top} \nabla^2 f(\bm{\theta}_t) \hat{\mathbf{v}}_t|\right)^{-1}}\mathbf{z}_t
\end{align}
This reduces the standard deviation of the noise compared to Eq. \eqref{eq:PreCon-FMALA}. We believe this helps contribute to poorer performance in higher dimensions as well as the aforementioned potential challenge of a mismatched covariance between the forward and reverse directions.

\subsubsection{PreCon-Line-FMALA}
For the proposal mechanism of PreCon-Line-FMALA,
\begin{align}
    \bm{\theta}_* &= \bm{\theta} + \bigg( \frac{\eta^2 D \nabla f( \bm{\theta}_t) \cdot \hat{\mathbf{v}}}{2 D |\hat{\mathbf{v}}^{\top} \nabla^2 f(\bm{\theta}_t) \hat{\mathbf{v}}|}  + \frac{ \eta \sqrt{D} z_t}{\sqrt{D|\hat{\mathbf{v}}^{\top} \nabla^2 f(\bm{\theta}_t) \hat{\mathbf{v}}|}}\bigg)\hat{\mathbf{v}}, \notag \\
    &= \bm{\theta} + \bigg( \frac{\eta^2 \nabla f( \bm{\theta}_t) \cdot \hat{\mathbf{v}}}{2 |\hat{\mathbf{v}}^{\top} \nabla^2 f(\bm{\theta}_t) \hat{\mathbf{v}}|}  + \frac{\eta z_t}{\sqrt{|\hat{\mathbf{v}}^{\top} \nabla^2 f(\bm{\theta}_t) \hat{\mathbf{v}}|}}\bigg)\hat{\mathbf{v}},
\end{align}
we recover the exact formulation of Eq. \eqref{eq:precon_line_FMALA_reduced}. Therefore the mechanism is unchanged.

\section{Additional Results}\label{app:exp}
\subsection{Funnel Distribution}\label{app:fun}

\subsubsection{Target Distribution}
We consider the funnel distribution introduced by \citet{neal2003slice}, defined as:
\[
\pi (\bm{\theta}, w) = \prod_{i=1}^D \mathcal{N}(\theta_i \mid 0, \exp(-w)) \cdot \mathcal{N}(w \mid 0, 3^2),
\]
where the coupling between $\bm{\theta}$ and $w$ creates a challenging geometry for MCMC samplers.

\subsubsection{Evaluation Metric: KL Divergence}
To evaluate performance, we compute the KL divergence between the true marginal $p(w) = \mathcal{N}(0, 3^2)$ and the empirical marginal $q(w)$ obtained from the samples. Specifically, we approximate $q(w)$ as a Gaussian using the empirical first and second moments from the samples $\{\tilde{w}_i\}_{i=1}^T$, and compute:
\[
\mathrm{D}_{\mathrm{KL}}(p(w) \Vert q(w)) = \frac{1}{2} \left( \frac{\sigma_q^2}{\sigma_p^2} + \frac{(\mu_q - \mu_p)^2}{\sigma_p^2} - 1 + \log\frac{\sigma_p^2}{\sigma_q^2} \right),
\]
where $(\mu_q, \sigma_q^2)$ are the empirical mean and variance of the $w$ samples.

\subsubsection{Experimental Setup}
Each experiment consists of 5 parallel MCMC chains, each with $10{,}000$ samples. Chains are independently initialized from a Gaussian distribution with zero mean and standard deviation 0.1. For each sampler and funnel dimension ($10D$, $50D$, $100D$), we perform a grid search over the step size $\eta$. At each grid point, we run 10 independent trials using different random seeds to assess variability and robustness. For each experiment we evaluate 100 grid points, which are evenly spread logarithmically from $0.1$ – $2.0$ for $D = 10$ and $50$, and from $0.01$ – $2.0$ for $D=100$. All these points and their standard deviation across the random seeds are shown in Figure \ref{fig:all_funnel_rows}.

\subsubsection{Statistical Testing}
We report the best KL divergence (i.e., lowest value) across grid points for each sampler. To assess whether observed differences are statistically significant, we apply a two-sample Welch’s $t$-test between the best-performing KL scores of each sampler and those of MALA. The resulting $p$-values are reported in Table~\ref{tab:funnel_data}.

\subsubsection{Effective Sample Size}
We compute the effective sample size (ESS) for each run using the \texttt{arviz.ess} implementation from \citet{arviz_2019}. We report the ESS separately for the $w$ variable and the average ESS across all $\bm{\theta}_i$ components. We include standard deviations over the 10 random seeds for all reported metrics.

\subsubsection{Step Size Sensitivity}
\begin{figure}[hbt!]
    \centering
    \includegraphics[width=.85\linewidth]{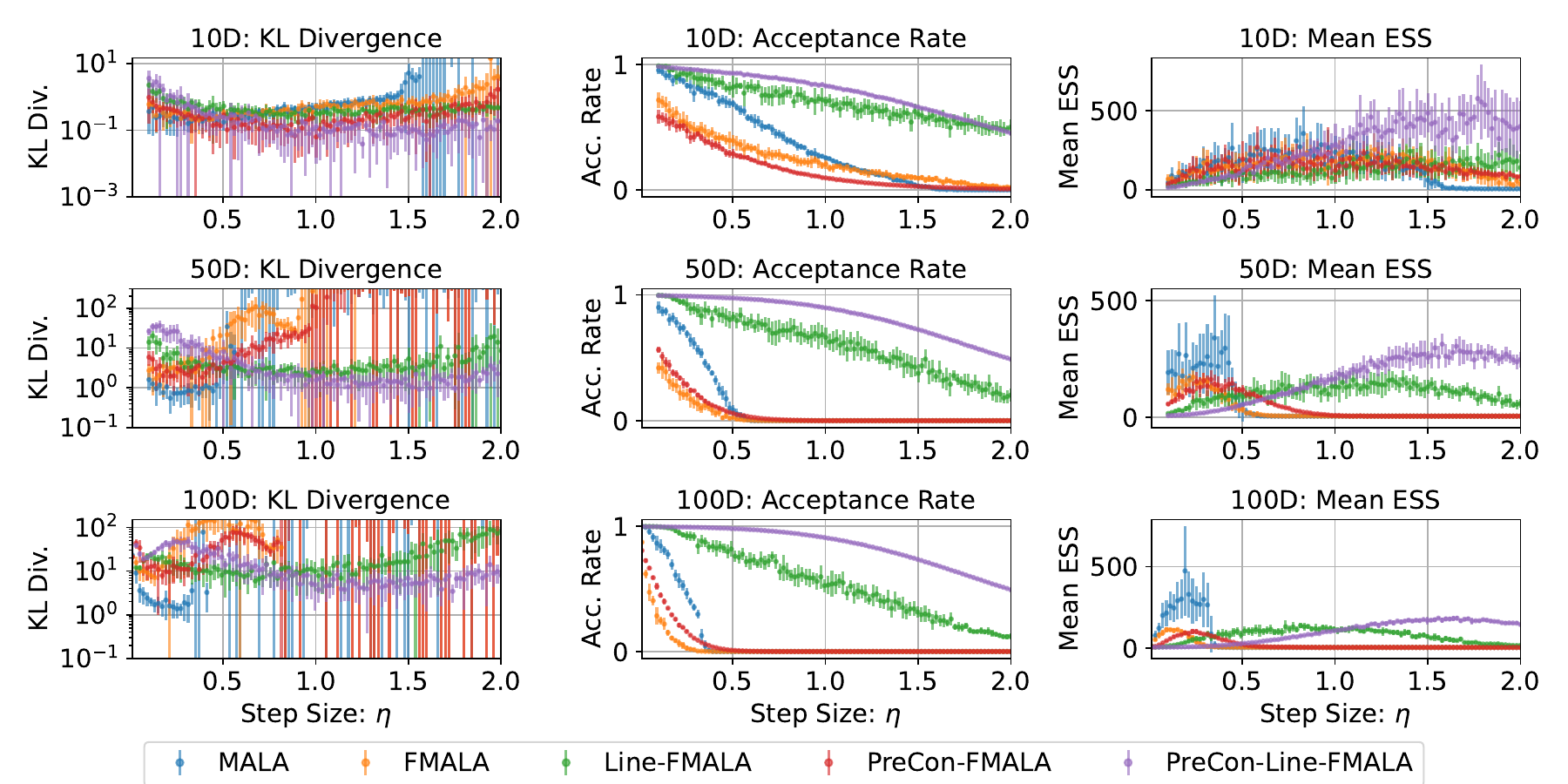}
    \caption{Full grid search corresponding to Table \ref{tab:funnel_data}.}
    \label{fig:all_funnel_rows}
\end{figure}

Figure~\ref{fig:10D_funnel} shows the KL divergence and acceptance rate as functions of the step size $\eta$ for the $10D$ funnel. Similar patterns are observed in higher dimensions. Notably, the line-based samplers (Line-FMALA and PC-Line-FMALA) exhibit broad stability across a wide range of $\eta$, while MALA shows sharp sensitivity—particularly in $100D$, where a small deviation from the optimal step size leads to near-zero acceptance (Figure~\ref{fig:all_funnel_rows}).

\subsection{Multinomial Logistic Regression}\label{app:blg}
\subsubsection{Target Distribution}
Our multinomial logistic regression model uses a Gaussian prior, $p(\bm{\theta}) = \prod_{d=1}^{D} \mathcal{N}(\bm{\theta}_d \mid \mathbf{0}, \mathbf{I})$, and likelihood 
$$p(\mathbf{y} \mid \mathbf{X}, \bm{\theta}= \{\mathbf{W}, \mathbf{b}\})
= \prod_{i=1}^{N} 
\frac{\exp\!\bigl(\mathbf{x}_i^\mathsf{T}\mathbf{W}_{y_i} \;+\; b_{y_i}\bigr)}
 {\sum_{c=1}^{C}\exp\!\bigl(\mathbf{x}_i^\mathsf{T}\mathbf{W}_{c} \;+\; b_{c}\bigr)},$$
 where $\mathbf{W}\in\mathbb{R}^{D_{\mathrm{in}}\times C}$,  $\mathbf{b}\in\mathbb{R}^C$, and $\mathbf{x}_i\in\mathbb{R}^{D_{\mathrm{in}}}$ for $N$ data.

\subsection{BNN for Regression} \label{app:bnn_regression}

\subsubsection{Model and Dataset}
We use a 5-layer fully connected neural network with hidden layers of 100 units and ReLU activations, resulting in $D=40{,}701$ parameters. The prior is Gaussian: $p(\bm{\theta}) = \prod_{d=1}^{D} \mathcal{N}(\bm{\theta}_d \mid \mathbf{0}, \sigma_{\mathrm{prior}}^{2})$ with $\sigma_{\mathrm{prior}} = 0.1$. The likelihood is $p(\mathbf{Y} \mid \mathbf{X}, \bm{\theta}) = \mathcal{N}(\mathbf{Y} \mid \mathrm{nn}(\mathbf{X};\bm{\theta}), \sigma_{\mathrm{lik}}^2)$ with $\sigma_{\mathrm{lik}} = 0.025$. We use the 400-point synthetic regression dataset from \citet{izmailov2020subspace}.

\subsubsection{Sampling Procedure}
Each sampler is run for $5\times10^4$ iterations. We discard the first $10^4$ samples as burn-in and thin by keeping every 100th sample. We perform Bayesian optimization over the step size before full runs. We note that for MALA, we further analyzed the results of the Bayesian optimization, and tested multiple further step sizes around the suggested optimal value. We found that its sensitivity to the step size for the budget of $5\times10^4$ samples meant that the step size was either too large resulting in a high rejection rate, or the step size was too small and it took too long to mix. We additionally optimized for the burn amount for MALA, whereas we stuck with the $10^4$ for all of the forward-mode samplers.

\subsection{Bayesian CNN}\label{app:cnn}

The architecture of the CNN is as follows. It consists of three convolutional blocks, where each block contains two instances of the following combination: 2D convolutional layer, followed by a ReLU. Each block finishes with a 2D max pooling layer with a $2\times2$ kernel. The first block goes from 3 to 32 channels, then second block goes from 32 to 64 channels, while the third block goes from 64 to 128 channels. All CNN layers use a kernel size of $3\times3$ and a padding of 1. After the three convolutional blocks, the model contains two linear layers. The exact details can be found in the attached code. 

Table \ref{tab:step_time_combined} includes the wall-clock timing results for the CNN with % JAX: 2395434
$2{,}396{,}330$ parameters. Unlike for Section \ref{sec:cnn}, we run these timing results ranging from the first $10{,}000$ images to the first $50{,}000$ images of CIFAR-10. We observed that MALA was unable to run through the initial JAX warmup-step (see \cite{jax2018github}) without reporting the RESOURCE EXHAUSTED error. We include this error below. This result is nice because it explicitly shows the increased memory cost associated with the reverse-mode sampler. 

\begin{table}[h!]
    \centering
    \caption{Time per step (s) for different training data sizes.}
    \label{tab:step_time_combined}
    \begin{tabular}{lccc}
        \toprule
        \textbf{Algorithm} & \bfseries {$N=10{,}000$} & \bfseries {$N=30{,}000$} & \bfseries {$N=50{,}000$} \\
        & \textbf{Time/Step (s)} & \textbf{Time/Step (s)} & \textbf{Time/Step (s)} \\
        \midrule
        FMALA            & 0.3979 $\pm$ 0.0008 & 0.9645 $\pm$ 0.0013 & 1.5475 $\pm$ 0.0054 \\
        Line-FMALA       & 0.3417 $\pm$ 0.0006 & 0.9089 $\pm$ 0.0025 & 1.4834 $\pm$ 0.0031 \\
        PC-FMALA         & 0.5283 $\pm$ 0.0021 & 1.3480 $\pm$ 0.0012 & 2.1750 $\pm$ 0.0019 \\
        PC-Line-FMALA    & 0.4706 $\pm$ 0.0009 & 1.2877 $\pm$ 0.0017 & 2.0982 $\pm$ 0.0033 \\
        MALA             & 0.5160 $\pm$ 0.0025 & Out-Of-Memory Error & Out-Of-Memory Error \\
        \bottomrule
    \end{tabular}
\end{table}

\begin{table}[h!]
    \centering
    \caption{Best test accuracy and corresponding calibration error on the first $10{,}000$ images of CIFAR-10 data. The full grid search is shown in Figure \ref{fig:cifar_pretrain} in the main paper.}
    \label{tab:cifar10_accuracy_ece}
    \begin{tabular}{lcc}
        \toprule
        \textbf{Algorithm} & \textbf{Accuracy} & \textbf{ECE} \\
        \midrule
        FMALA         & 0.7977 $\pm$ 0.0022 & 0.0250 $\pm$ 0.0014 \\
        L-FMALA       & 0.7971 $\pm$ 0.0024 & 0.0287 $\pm$ 0.0031 \\
        PC-L-FMALA    & 0.7964 $\pm$ 0.0019 & 0.0285 $\pm$ 0.0019 \\
        MALA          & 0.7979 $\pm$ 0.0022 & 0.0263 $\pm$ 0.0026 \\
        \midrule
        Pre-trained via SGD           & 0.7942 & 0.0360 \\
        \bottomrule
    \end{tabular}
\end{table}

\begin{tcblisting}{title={Python Error\\\small\bfseries NVIDIA RTX A6000 48 GB, CUDA 12.8, JAX 0.5.0},
  colback=black!5, colframe=red!75!black,
  fonttitle=\bfseries,
  listing only,
  breakable,
  listing options={
    basicstyle=\ttfamily\small,
    columns=fullflexible,
    breaklines=true
  }}
...
The above exception was the direct cause of the following exception:

Traceback (most recent call last):
  File "/homes/usr/lib/scripts/Sampling/cifar10_cnn/cifar10_time.py", line 452, in <module>
    main()
  File "/homes/usr/lib/scripts/Sampling/cifar10_cnn/cifar10_time.py", line 390, in main
    dummy_theta, dummy_accepted = sampler_step_jitted(state, rng_key, epsilon)
jaxlib.xla_extension.XlaRuntimeError: RESOURCE_EXHAUSTED: Out of memory while trying to allocate 58388679680 bytes.
\end{tcblisting}

%%%%%%%%%%%%%%%%%%%%%%%%%%%%%%%%%%%%%%%%%%%%%%%%%%%%%%%%%%%%

\end{document}